\documentclass{article}

\usepackage[utf8]{inputenc}
\usepackage[T1]{fontenc}
\usepackage{lmodern}
\usepackage{amsmath,amssymb, amsthm}
\usepackage{enumitem}
\usepackage[hidelinks]{hyperref}
\usepackage[margin=2cm]{geometry}
\usepackage{graphicx}
\usepackage[authoryear]{natbib}
\usepackage{mathtools}
\usepackage{colortbl}
\usepackage{booktabs}    
\usepackage{algorithm}
\usepackage{algpseudocode}
\usepackage{bm}
\usepackage{accents} 
\usepackage{authblk}
\usepackage{xcolor}
\usepackage{footmisc}

\newtheorem{fact}{Fact}
\newcommand{\ABA}{{\sf ABA}}
\newcommand{\METIS}{{\sf METIS}}
\newcommand{\NA}{---}
\newcommand{\floor}[1]{\left\lfloor #1 \right\rfloor}
\newcommand{\ceil}[1]{\left\lceil #1 \right\rceil}

\newtheorem{lemma}{Lemma}

\title{A Fast and Effective Method for Euclidean Anticlustering: The Assignment-Based-Anticlustering Algorithm}

\date{}

\author[1]{P.~Baumann\thanks{Corresponding author: philipp.baumann@unibe.ch}}
\author[2]{O.~Goldschmidt}
\author[3]{D.~S.~Hochbaum}
\author[3]{J.~Yang}

\affil[1]{Department of Business Administration, University of Bern, Engehaldenstr.\ 4, 3012 Bern, Switzerland}
\affil[2]{Riverside County Office of Education, Riverside, CA 92501, USA}
\affil[3]{Industrial Engineering and Operations Research Department, University of California, Berkeley, CA 94720, USA}
\affil[ ]{\texttt{philipp.baumann@unibe.ch, goldoliv@gmail.com, dhochbaum@berkeley.edu, jason\_yang@berkeley.edu}} 

\begin{document}

\maketitle
\begin{abstract}
Anticlustering is an NP-hard combinatorial optimization problem that consists of partitioning a set of objects into equal-sized groups called anticlusters such that the objects in the same anticluster are as dissimilar as possible and thereby representative of the entire set of objects. Here we study the case where the dissimilarity metric is the squared Euclidean distance between the respective feature vectors. Applications of Euclidean anticlustering include social studies, cross-validation, creating mini-batches for stochastic gradient descent, and finding balanced $K$-cut partitions. In particular, machine-learning applications such as mini-batch generation involve million-scale datasets and very large values of $K$, making scalable anticlustering algorithms essential. We propose a new algorithm, the Assignment-Based Anticlustering (\ABA) algorithm, that scales to instances with millions of objects and hundreds of thousands of anticlusters within seconds to minutes, which is far beyond what existing anticlustering methods can manage. We demonstrate here, via an extensive computational study, that our algorithm outperforms existing anticlustering methods in both solution quality and running time. This is so also for anticlustering with categories. For the related problem of balanced $K$-cut partitioning, our algorithm is superior to the well-known \METIS~method. The code of our algorithm is available on GitHub. 
\end{abstract}

\section{Introduction}\label{sec:introduction}

Anticlustering is an NP-hard combinatorial optimization problem that consists of partitioning a set of objects into equal-sized groups called anticlusters such that the sum of distance weights between pairs of objects in the same anticluster is maximized. 
This is with the goal of achieving anticlusters whose objects are as dissimilar as possible and thereby representative of the entire set of objects. Achieving diversity within groups is the focus of applications from various fields including operations research (\citealt{weitz1997empirical, baker2002methods,palubeckis2015maximally,lai2016iterated,schulz2021balanced,schulz2023balanced_a, schulz2023balanced_b,hochbaum2023breakpoints}), management science (\citealt{krass2006university}), psychology and social sciences (\citealt{brusco2020combining, papenberg2021using}), bioinformatics (\citealt{papenberg2025anticlustering}), VLSI circuit design (\citealt{chen1986placement,feo1990class}), computer systems and memory management (\citealt{kral1965problem}), and more recently machine learning (\citealt{mauri2023robust}). 

Several anticlustering variants have been proposed in the literature that differ in a number of ways: how the objects are represented, as nodes in an edge-weighted graph or as feature vectors; how the distance weights are measured, as distance metrics or arbitrary weights; 
which objective is optimized, the most common objective being maximizing within-anticluster sum of weights, or alternatively maximizing the minimum distance weight between any two objects in the same anticluster; 
and which alternative constraints are considered, requiring each anticluster's size to satisfy some lower and upper bounds instead of equal-size anticlusters or additional constraints for the case when objects are associated with categories, that ensure that each anticluster has the same number of objects of a given category. 

The anticlustering problem has been studied under the name the {\em balanced $K$-cut} problem. The balanced $K$-cut problem consists of partitioning the nodes of a graph into $K$ anticlusters each of which has at most $\ceil{N/K}$ nodes, where $N$ denotes the number of nodes in the graph. The objective is to minimize the sum of edge weights between the anticlusters (the cut cost). 
This objective function of the balanced cut problem is the complement of the anticlustering objective, since the weight of the edges in the cut is the total sum of weights of all edges minus the sum of the edge weights inside the anticlusters, which is the objective of the anticlustering problem, and thereby equivalent. The anticlustering problem was also studied under the name maximally diverse grouping problem (MDGP), e.g.\ \citealt{gallego2013tabu}. 

We focus here on anticlustering in Euclidean spaces, where objects are represented as feature vectors and distance weights are measured using squared Euclidean distances. The objective is to assign an equal number of objects to each anticluster so that the sum of pairwise squared Euclidean distances between objects within anticlusters is maximized. Euclidean anticlustering can be traced back to \cite{spath1986anticlustering} and is now arguably the most widely-used variant of anticlustering in practice. Euclidean anticlustering is used in psychology and social sciences to create participant groups that are comparable and representative of the entire dataset (\citealt{papenberg2021using}). In biomedical research such as high-throughput sequencing, it is used to assign patient DNA and RNA samples to batches to limit batch effects (see \citealt{papenberg2025anticlustering}). 
In unsupervised learning, it is used to split datasets into representative subsets to help compute lower bounds for instances of the minimum sum-of-squares clustering problem (see \citealt{croella2025strong}). In supervised learning, it is used in cross-validation to create representative folds that support reliable model evaluation (see \citealt{papenberg2021using}), and in neural network training to form mini-batches for stochastic gradient descent, improving model generalization and convergence (see \citealt{joseph2019submodular, wang2019fixing, banerjee2021deterministic}). These recent applications in machine learning involve very large instances of Euclidean anticlustering with millions of objects and thousands of anticlusters. Although anticlustering is often considered easier on large datasets because random partitions are more likely to produce reasonable solutions (see \citealt{papenberg2021using}), this holds only when the number of anticlusters is small. The deterioration of the quality of random partitions with increasing number of anticlusters is clearly apparent in the empirical study here. Therefore random partitions, which are used in practice for applications such as mini-batch generation for neural network training, often provide low-quality solutions, creating a need for alternative anticlustering algorithms.

Various exact and heuristic methods have been developed for anticlustering. Exact methods are primarily based on mixed-integer programming formulations and are limited to small instances, solving problems with up to about 100 objects \citep{gallego2013tabu,papenberg2021using,schulz2021balanced,schulz2022new,schulz2023balanced_b}. 
For larger instances, numerous heuristics have been introduced. \cite{feo1992one} propose approximation algorithms for instances in which anticlusters contain exactly three or four objects. These algorithms run in cubic time in the number of objects and are therefore not suitable for large-scale applications. For metric spaces, \cite{hassin2003approximation, hassin2006improved} propose approximation algorithms for anticlustering with prescribed, not necessarily equal, anticluster sizes. These algorithms have improved running times compared to the algorithms of \cite{feo1992one}. Since these algorithms require all pairwise distances between objects as an input, when applied to Euclidean anticlustering, this limits their scalability. 
Other heuristics which require the complete distance matrix as input when used for Euclidean anticlustering face the same scalability issue (see \citealt{palubeckis2015maximally,lai2016iterated, singh2019new, yang2022three}). 
Several greedy constructive methods for anticlustering have been proposed in the context of mini-batch generation for training deep neural networks (see \citealt{joseph2019submodular,wang2019fixing,banerjee2021deterministic}). These methods generate mini-batches that enable faster convergence and higher neural network accuracy, however, they remain computationally expensive, requiring hours of running time for processing standard image datasets such as ImageNet. 
Many state-of-the-art anticlustering heuristics are {\em exchange-based} (see \citealt{papenberg2021using}). Such methods start from a random partition and iteratively improve it by exchanging pairs of objects from different anticlusters. The default method is to select an object and evaluate all pairwise exchanges between the selected object and objects from other anticlusters. The exchange that achieves the largest improvement in the objective value is performed. The process terminates when the exchange process has been repeated for each object. This default method is computationally prohibitive and the most-scalable methods restrict the number or the type of permitted exchanges. 
The leading exchange-based methods are available in the R package called \texttt{anticlust} (see \citealt{papenberg2021using}).

We introduce here a novel heuristic for Euclidean anticlustering, the Assignment-Based Anticlustering (\ABA) algorithm. The \ABA~algorithm is a constructive heuristic that builds a solution with a single pass over the objects. It solves a series of assignment problems that assign batches of objects to anticlusters. The batches are generated from a ranked sequence of objects in terms of their distance to the global centroid of the dataset, where the {\em centroid} of a set of vectors is the vector of the average entries. The sequence is broken down into subsequences (batches) of the first $K$ objects, the next $K$ objects, and so forth. The objects of the first batch are the initial singleton anticlusters. 
To assign the $K$ objects from the next batch, we solve the maximum weight assignment of the $K$ objects to the $K$ incumbent anticlusters. 
The weights are the distances (squared Euclidean) between the objects in the batch and the centroids of the incumbent anticlusters. After each batch assignment, the centroids of anticlusters are updated. 
Note that the last batch could be of size less than $K$. 

Some features of \ABA\ make it very efficient. Firstly, each object is assigned only once to its anticluster. Secondly, \ABA\ computes the distance of each object in a batch to the centroid of the candidate anticlusters, instead of computing the distances to all objects in the anticlusters.
\ABA\ also delivers solutions with desirable statistical properties. One desirable property is {\em anticluster similarity}, i.e., that the anticlusters are similar to each other in terms of mean and variance of the distances within the anticlusters (see \citealt{schulz2021balanced, papenberg2024k}).
The assignments of \ABA\ guarantee that there is exactly one object of each batch in each resulting anticluster, except maybe for the last batch. 
For {\em centrality} defined as the distance to the global centroid of the dataset, we note that the objects in each batch have similar centralities. 
\ABA\ therefore delivers solutions in which every anticluster has a representation of the entire range of centralities. 
\ABA~is shown in our experimental study to significantly outperform the other methods in terms of this anticluster similarity.




In addition to the base variant of \ABA\ described above, we introduce here two other variants: \ABA\ for anticlustering with categories and \ABA\ with a hierarchical decomposition strategy. 
Anticlustering with categories is an extension of the anticlustering problem  where objects are associated with categories and each category must be equally represented in every anticluster (see \citealt{croella2025strong, papenberg2021using}). \ABA\ with a hierarchical decomposition strategy is a variant designed for very large-scale instances, which achieves drastic speedups with minor effect on the quality of the solution. 

For the special case of anticlusters of size two, the Euclidean anticlustering problem is equivalent to the Euclidean maximum-weight non-bipartite matching problem. 
For this case, the work in \cite{baumann2026fast} introduces a variant of \ABA, \ABA-matching, that is based on different sequencing of the objects. 
This variant was shown in \cite{baumann2026fast} to outperform state-of-the-art algorithms for non-bipartite matching in terms of the trade-off between running time and quality of solution. 
Compared to exact algorithms, \ABA-matching delivers almost optimal solutions, at worst $1.7\%$ off, while offering very fast running times, whereas exact algorithms fail to deliver a solution within one hour of running time for the larger datasets. Also compared to heuristics, \ABA-matching delivers consistently better quality of solutions and much faster running times.

To summarize, our contributions here are: 

\begin{enumerate}
    \item The introduction of the highly-scalable assignment-based algorithm (\ABA) for the anticlustering problem. 
    \item Devising a variant of \ABA\ for anticlustering with categories, which achieves superior results compared to leading methods of \cite{croella2025strong} and \cite{papenberg2021using}.
    \item For very large-scale datasets and very large values of $K$ (e.g., $K=100{,}000$), we propose a variant of \ABA\ that employs a hierarchical decomposition strategy resulting in drastically reduced running times while maintaining high solution quality.   
    \item We provide the most comprehensive computational study of anticlustering algorithms to date, using a broad spectrum of published datasets.  
    \item It is demonstrated empirically that the \ABA~algorithm outperforms the leading exchange-based anticlustering methods in both solution quality and running time. Unlike existing methods, \ABA\ also scales to very large instances with millions of objects and thousands of anticlusters within seconds to minutes.
    \item We generate new insights about the performance of random partitioning, that its performance deteriorates with smaller size anticlusters. This poor performance trend applies also to all benchmark algorithms tested here.     
    \item For the extensively studied anticlustering problem under the name {\em balanced $K$-cut}, \ABA~is shown here to be very effective for the Euclidean case. In particular, it achieves superior results compared to the state-of-the-art \METIS\ method in terms of both solution quality and running time.
    \item It is shown that the design of \ABA\ is such that it not only delivers on the anticlustering objective, but also on the anticluster similarity objective. In all experiments, \ABA\ consistently delivers substantially higher anticluster similarity than the benchmark algorithms.  
\end{enumerate}


The remainder of the paper is organized as follows. Section~\ref{sec:problem} formalizes the anticlustering problem and introduces the notation used throughout. Section~\ref{sec:literature} reviews existing algorithms for Euclidean anticlustering and related problems. Section~\ref{sec:algorithm} presents the Assignment-Based Anticlustering (\ABA) algorithm, including its base variant; anticlustering with categories; the hierarchical decomposition strategy; and complexity analysis. 
Section~\ref{sec:experiment} reports our experimental results. Section~\ref{sec:conclusions} concludes and outlines future research directions.

\section{Problem Description}\label{sec:problem}
Let $G=(V, E)$ be an undirected, weighted graph with $V=\{1, \ldots, N\}$ the set of nodes (objects), $E \subseteq \{ \{i, i'\}: i < i', i, i' \in V\}$ the set of edges, and edge weights $w_{ii'}\geq0$ representing dissimilarities between nodes $i$ and $i'$. The anticlustering problem consists of partitioning the set of nodes into $K$ groups (anticlusters) of equal size such that the sum of edge weights within each group is maximized (see \citealt{spath1986anticlustering}). The problem is NP-hard because even the case of partitioning into two equal-sized groups is at least as hard as the balanced cut problem (see \citealt{garey1990computers}). In this work, we focus on the most common variant, where objects are points in a $D$-dimensional Euclidean space and pairwise dissimilarities are defined by squared Euclidean distances.

Formally, the anticlustering problem we study can be described as follows. Given a dataset of $N$ objects ${\mathcal{X}}=\{\bm{x}_1, \ldots, \bm{x}_N\}$ with $D$ features, where $\bm{x}_i=[x_{i1}, \ldots, x_{iD}] \in \mathbb{R}^D$ is the $i$-th object, and $x_{id}$ is the value of the $d$-th feature of this $i$-th object. Let ${\mathcal{N}} = \{1,\ldots, N\}$ denote the set of object indices. 
The problem is to find a partition $\mathcal{C}=\{\mathcal{C}_1, \ldots, \mathcal{C}_K\}$ of the objects into $K$ anticlusters of ``approximately'' equal size such that the total sum of squared Euclidean distances between objects within the same anticluster is maximized. Here we interpret ``approximately'' equal size as sizes that are strictly less than one unit away from the average size ($N/K$). Let $\mathcal{K} = \{1, \ldots, K\}$ denote the set of anticluster indices, $\mathcal{C}_k$ the set of object indices in anticluster $k$, and $d(i, i') = \|\bm{x}_i - \bm{x}_{i'}\|_2^2$ the squared Euclidean distance between objects $i$ and $i'$. With this notation, the Euclidean anticlustering problem (EA) can be formulated as follows:

\[
\text{(EA)}\;
\left\{
\begin{aligned}
\max_{\mathcal{C}} \quad &
W(\mathcal{C}) = \sum_{k \in \mathcal{K}} \sum_{i,i' \in \mathcal{C}_k,\, i<i'} d(i,i') & (1) \\[4pt]
\text{s.t.} \quad &
\lfloor N / K \rfloor \le |\mathcal{C}_k| \le \lceil N / K \rceil \quad (k \in \mathcal{K}) & (2) \\[4pt]
&
\bigcup_{k \in \mathcal{K}} \mathcal{C}_k = \mathcal{N} & (3) \\[4pt]
&
\mathcal{C}_k \cap \mathcal{C}_{k'} = \emptyset \quad (k, k' \in \mathcal{K},\, k < k') & (4)
\end{aligned}
\right.
\]

The objective function~(1) is the sum of pairwise squared Euclidean distances between objects assigned to the same anticluster. We refer to the objective function value as the diversity $W(\mathcal{C})$ of the anticlustering solution $\mathcal{C}$. Constraints~(2) ensure that the size of two anticlusters differ by at most one. Constraints~(3) and (4) ensure that every object is assigned to exactly one anticluster. 

The Euclidean anticlustering problem with categories (EAc) is
an extension of the Euclidean anticlustering where each object is associated with a category $g$ from a given set of categories $\mathcal{G}=\{1, \ldots, G\}$. 
Let $\mathcal{N}_g = \{i \in \mathcal{N}| \text{category}(i) = g\}$ be the indices of the objects that belong to category $g \in \mathcal{G}$. With this additional notation, the Euclidean anticlustering problem with categories (EAc) is formulated as follows: 

\[
\text{(EAc)}\;
\left\{
\begin{aligned}
\max_{\mathcal{C}} \quad & 
W(\mathcal{C}) \\[4pt]
\text{s.t.}\quad & (2), (3), (4) \\[4pt]
& \displaystyle \left\lfloor \frac{|\mathcal{N}_g|}{K} \right\rfloor \le \big|\mathcal{C}_k \cap \mathcal{N}_g\big| \quad (k\in\mathcal{K}, g\in\mathcal{G}) & (5)\\
& \displaystyle \big|\mathcal{C}_k \cap \mathcal{N}_g\big| \le \left\lceil \frac{|\mathcal{N}_g|}{K} \right\rceil \quad (k\in\mathcal{K}, g\in\mathcal{G}) & (6) \\
\end{aligned}
\right.
\label{form:A_categories}
\]

In this extended formulation, constraints~(5) and (6) ensure that for each category $g \in \mathcal{G}$, the number of objects belonging to category $g$ differs by at most one object between each pair of anticlusters. Specifically, the number of objects from category $g$ assigned to any anticluster $\mathcal{C}_k$ is constrained to lie between 
$\left\lfloor |\mathcal{N}_g| / K \right\rfloor$ and $\left\lceil |\mathcal{N}_g| / K \right\rceil$, so that all anticlusters represent the overall category distribution. 

\section{Literature Review}\label{sec:literature}

In this section, we review existing algorithms for the Euclidean anticlustering or closely related problems. Section~\ref{sec:literature:exact} discusses exact approaches and Section~\ref{sec:literature:heuristic} reviews heuristic approaches. 

\subsection{Exact Approaches}\label{sec:literature:exact}

Although there are no exact approaches developed specifically for Euclidean anticlustering, the general anticlustering methods obviously apply for that case. \cite{gallego2013tabu} used a quadratic integer programming formulation for the anticlustering problem and a variant allowing for the size of each anticluster to be constrained by an upper and lower bound.    
In their paper, the authors refer to the anticlustering problem under the name {\it maximally diverse grouping problem} (MDGP). 

\cite{grotschel1989cutting} introduced a {\it linear} integer programming formulation for a related clustering problem. This formulation was extended and applied to anticlustering by \cite{papenberg2021using}.

\cite{schulz2021balanced} studied the anticlustering problem with few features in which dissimilarities are computed using the Manhattan distance ($L_1$). For special cases, such as when $N/K \leq 3$ and a single feature, the problem is shown to be polynomial-time solvable. For a single feature and non-negative integer feature values, the problem is proved to be weakly NP-hard. However, any slight generalization, including allowing the feature values to be rational, renders the problem to be strongly NP-hard. The author also presents an integer programming formulation that is tested for instances with one feature or two features. 
In a follow-up study, \cite{schulz2022new} compared a variant of this formulation to the formulations of \cite{gallego2013tabu} and \cite{papenberg2021using} and found that it performs best for $L_1$ distances. \cite{schulz2023balanced_b} extended this work and proposed a two-step exact mixed-integer programming approach for anticlustering with $L_1$ distances, reporting optimal solutions for instances on one feature, two features, and five features, with up to $200$ objects. 
\cite{schulz2023balanced_a} demonstrated that an enhanced integer programming formulation solves the problem with single feature, integer feature values, and $L_1$ distances, for instances with up to $6{,}000$ objects to optimality within reasonable computation times (less than 10 minutes). 

Exact approaches were also proposed for other variants of anticlustering. For example, \cite{papenberg2025extending} proposed an exact approach for an anticlustering problem where the goal is to maximize the minimum distance between any two objects within the same anticluster (maximum dispersion). The approach is based on multiple calls for solving the $K$-coloring problem. 
Another variant, introduced by \citealt{brusco2020combining}, is the bicriterion anticlustering problem. Bicriterion anticlustering is anticlustering that seeks to maximize the sum of distances within the anticlusters subject to the maximum dispersion exceeding a user-specified lower bound. 
\cite{papenberg2025extending} showed how an extension of the exact approach for maximum dispersion is effective also for the bicriterion anticlustering problem. 

All the exact approaches discussed above work only for small instances, which is far below the scale of practical instances. 

\subsection{Heuristic Approaches}\label{sec:literature:heuristic}

\cite{spath1986anticlustering} introduced Euclidean anticlustering where dissimilarity is measured by squared Euclidean distances with specified number of anticlusters but without anticluster-size constraints. The author introduced a move-based heuristic that moves a single object from one anticluster to another. At any iteration, this results in a feasible solution because there are no size constraints. \cite{spath1986anticlustering} also observed that random partitions do not work well. 

\cite{weitz1997empirical} proposed a heuristic for anticlustering, with equal-sized anticlusters, that considers pairwise exchanges of objects between anticlusters instead of single-object moves. The heuristic first randomly assigns objects to anticlusters, and then for one object at a time considers exchanges with all objects from other anticlusters. The exchange that improves the objective function the most is executed. Note that an exchange is only conducted if it improves the objective function. The exchange procedure is terminated when there is no improving exchange for all objects. 

\cite{papenberg2021using} presented two types of exchange-based heuristics. The first type is the exchange-based heuristic of \cite{weitz1997empirical} which uses by default a different stopping criterion, i.e., it stops after a single pass of exchanges through all objects. The second type is an exchange-based heuristic called \texttt{fast-anticlustering} that is specifically designed for large instances of Euclidean anticlustering. It works in the same way as the standard exchange-based heuristic, but considers only a user-defined number of exchange partners for each object and uses a different formulation of the objective function, which only considers the distances between the anticluster centroids and the global centroid instead of all distances between objects and the centroid of their assigned anticluster. The default behavior of \texttt{fast-anticlustering} is to generate exchange partners by using a nearest neighbor search. Using more exchange partners tends to improve the quality of the solution, but increases running time. To avoid running the nearest neighbor search, the exchange partners can be chosen randomly. Selecting exchange partners randomly is significantly faster. An advantage of exchange-based heuristics is that additional constraints can easily be integrated into the exchange process. As an example, exchange-based heuristics can be applied with minor modifications to Euclidean anticlustering with categories (see Section~\ref{sec:problem}), and anticlustering with must-link and cannot-link constraints between pairs of objects.

As noted by \cite{papenberg2024k} the definition of anticlustering does not fully capture the needs in practical applications. The objective of anticlustering encourages anticlusters that have centroids very close to the global centroid. For instance, two anticlusters that have objects arranged along concentric circles of very different radii would be an optimal solution for an anticlustering problem. However, their statistical distribution properties are very different. In most practical applications, the anticlusters should have similar statistical properties. That is, the sum of pairwise distances within all anticlusters should be approximately equal - achieving anticluster similarity. This secondary objective of anticluster similarity was addressed by \cite{papenberg2024k} for tabular input data. They proposed to use the exchange-based heuristic with adjusted input data that involves adding, for each feature, a list of statistical-moment features, e.g., variance, skewness, or kurtosis. Adding $p$ additional statistical moments increases the dimension of the original feature space by a factor of $p + 1$. This strategy's significant increase of the dimensionality of the data leads to very high computational costs. 

Numerous heuristics have been developed for anticlustering under the name maximally diverse grouping problem (MDGP): \cite{gallego2013tabu, palubeckis2015maximally,lai2016iterated,singh2019new,lai2021neighborhood}, and \cite{yang2022three}. These heuristics are mainly based on metaheuristic frameworks such as tabu search, iterated local search, variable neighborhood search, and genetic algorithms. Among those metaheuristics, the method of \cite{yang2022three} is reported to dominate the others in this group. 
This method of \cite{yang2022three} is a population-based metaheuristic that is included in the R-package \texttt{anticlust}.  
For the anticlustering with Manhattan distance problem, \cite{schulz2023balanced_b} presented exact approaches, complexity study, and a number of heuristics. These heuristics are based on the case of a single feature that is then extended to account for multiple features.  


Several approximation algorithms are applicable to Euclidean anticlustering. \cite{feo1992one} proposed one-half approximation algorithms for special cases of the $K$-partition problem in which all partitions have size three ($K=N/3$) or four ($K=N/4$). \cite{hassin2003approximation} proposed an approximation algorithm for the problem with prescribed, not necessarily equal, anticluster sizes and distances that satisfy the the triangular inequality. Their algorithm offers improved running time compared to the algorithm of \cite{feo1992one} and obtains an approximation ratio of 0.375. \cite{hassin2006improved} introduced an improved approximation algorithm for the same problem, whose approximation ratio is $(\frac{1}{2} - \frac{3}{q})$, where $q$ is the size of the smallest anticluster. 


Other types of anticlustering heuristics have been developed in the context of mini-batch construction for machine learning. \cite{wang2019fixing} proposed a method to generate representative, equal-size mini-batches for training neural networks. Their approach uses a greedy algorithm that starts with empty mini-batches and iteratively selects the least representative one.
Representativeness of a set $S$ is measured as the sum of distances of all objects not in the set to their nearest object in set $S$. From the set of unassigned objects, the algorithm assigns the object that improves the representativeness of the selected mini-batch the most, repeating this process until all mini-batches reach the desired size. 
The authors further introduced a hierarchical variant of the algorithm to reduce memory and computational costs, and show 
for the datasets CIFAR-100 and ImageNet, that the resulting mini-batch sequences consistently outperform random partitioning. However, the overall running time of the approach is dominated by the cost of constructing the distance matrix. This is reported, in \cite{wang2019fixing}, to require approximately $3$ minutes for CIFAR-100 and almost $90$ minutes for ImageNet on a 16-core CPU.

\cite{banerjee2021deterministic} proposed another approach for generating mini-batches. Their approach is based on integer programming that is relaxed to linear programming, the solution of which is rounded. This approach is computationally expensive since the linear program must be solved for each mini-batch. 

The leading algorithm for anticlustering as balanced $K$-cut is the multilevel graph partitioning algorithm of \cite{karypis1998fast}, which is implemented in the well-known \METIS\ software. The algorithm reduces the size of the graph by collapsing vertices and edges, then partitions the smaller graph, and then ``uncoarsens" it to construct a partition for the original graph. 

\section{The New Assignment-Based Anticlustering (\ABA) Algorithm}\label{sec:algorithm}
The \ABA~algorithm constructs an anticlustering solution by iteratively solving an assignment problem of $K$ objects in a batch to the $K$ anticlusters. The key idea in creating the batches is first to produce a ranked list of the objects in decreasing distance to the global centroid and then to create $\ceil{N/K}$ batches formed by the subsequences: the first $K$ objects, the next $K$ objects, etc.\ with the last batch possibly including fewer than $K$ objects. Each object of the first batch forms an initial singleton anticluster. In each iteration, the algorithm assigns the next batch of $K$ objects (or possibly less for the last batch) and assigns it to anticlusters such that the total squared Euclidean distance between the objects and the centroids of the anticlusters is maximized. The centroids of the anticlusters are then updated based on the new assignments. With increasing number of 
iterations, this strategy tends to move the centroids of the anticlusters towards the global centroid of the data set. Since the final positions of the anticluster centroids tend to be close to the global centroid, the proposed batch selection strategy will usually result in anticlusters having similar distributions of distances between their objects and the respective anticluster centroid. 

By representing anticlusters through their centroids, the \ABA~algorithm takes advantage of the equivalence between maximizing the within-cluster sum of squares and the within-cluster variance (see Fact~\ref{fact1}). This allows distances to be computed between objects and centroids rather than between pairs of objects, resulting in greater computational efficiency. The following well-known fact, Fact~\ref{fact1}, establishes the equivalence between the sum of distances to the centroids of the respective anticlusters to the sum of squares of distances within the anticlusters. For completeness' sake, the proof is provided in the appendix.

\begin{fact}\label{fact1}
For anticluster $\mathcal{C}_k$ and centroid $\bm{\mu}_k = \frac{1}{|\mathcal{C}_k|}\sum_{i \in \mathcal{C}_k} \bm{x}_i$:
\[
\sum_{i,i' \in \mathcal{C}_k,\, i<i'} \|\bm{x}_i - \bm{x}_{i'}\|_2^2 
= |\mathcal{C}_k| \sum_{i \in \mathcal{C}_k} \| \bm{x}_i - \bm{\mu}_k \|_2^2.
\]
\end{fact}

In Section~\ref{sec:algorithm:base}, we first describe the base variant of the algorithm. In  
Section~\ref{sec:algorithm:categories}, we introduce a variant of the \ABA~algorithm that can be applied to the Euclidean anticlustering problem with categories (see Section~\ref{sec:problem}). In Section~\ref{sec:algorithm:hierarchical}, we present  a hierarchical decomposition strategy that substantially speeds up the \ABA~algorithm, in particular when $K$ is large. Finally, in Section~\ref{sec:algorithm:complexity}, we discuss the running time complexity of the \ABA~algorithm. 

\subsection{Base Variant of the \ABA~Algorithm}\label{sec:algorithm:base}
Algorithm~\ref{alg:base} provides the pseudocode of the \ABA~algorithm. The \ABA~algorithm takes as input a dataset ${\mathcal{X}}=\{\bm{x}_1, \ldots, \bm{x}_N\}$ consisting of $N$ objects, each represented by a $D$-dimensional feature vector, and a parameter $K$ which specifies the desired number of anticlusters. The algorithm starts by computing the squared Euclidean distances between each object $\bm{x}_i$ with $i=1,\ldots,N$ and the centroid of the dataset $\bm{\mu}=\frac{1}{N} \sum_{i=1,\ldots,N} \bm{x}_i$. The algorithm then creates a list $\mathcal{N}^{\downarrow}$ which contains the indices of the objects, sorted by their squared Euclidean distance to the global centroid $\bm{\mu}$ in descending order. The list $\mathcal{N}^{\downarrow}$ is split into $B=\lceil N / K \rceil$ batches. The first batch $\mathcal{B}_1$ contains the first $K$ indices of $\mathcal{N}^{\downarrow}$, the batch $\mathcal{B}_2$ the next $K$, and so on. Thus, all batches contain $K$ indices, with the possible exception of the last batch $\mathcal{B}_B$ if $N$ is not divisible by $K$. Next, the algorithm assigns each object whose index is contained in $\mathcal{B}_1$ to one of the $K$ anticlusters and initializes the centroid of each anticluster with the feature vector of its assigned object. The remaining $B-1$ batches are processed as follows. Given a batch $\mathcal{B}_b$, a cost matrix of size $|\mathcal{B}_b| \times K$ is computed where each entry $j, k$ $j=1, \ldots, |\mathcal{B}_b|$ and $k=1, \ldots, K$ corresponds to the squared Euclidean distance between the corresponding object $\bm{x}_{\mathcal{B}_b[j]}$ and the corresponding centroid $\bm{\mu}_k$ of anticluster $k$. The cost matrix is passed to a linear assignment problem solver that determines the maximal cost assignment. Based on the new assignments, the centroids of the anticlusters are updated. After the processing of all batches, the \ABA~algorithm returns a vector of length $N$ that contains for each object the label (index) of the anticluster to which it was assigned. 

\begin{algorithm}

\caption{Assignment-based Anticlustering (\ABA)}
\vskip6pt
\begin{algorithmic}
\Function{ABA}{${\mathcal{X}}=\{\bm{x}_1, \ldots, \bm{x}_N\}$, $K$}

	\State $\text{labels} \gets \text{empty vector of size } N$
	\State $\text{distances} \gets \text{empty vector of size } N$
	\State $\bm{\mu} \gets \frac{1}{N}\sum_{i \in \mathcal{N}}\bm{x}_i$
	\For{$i=1$ \textbf{to} $N$}
		\State $\text{distances}[i] \gets \|\bm{x}_i - \bm{\mu}\|_2^2$ 
	\EndFor
	\State $\mathcal{N}^{\downarrow} \gets \text{argsort}(\text{distances})$ \Comment{Sort objects in descending order and store their indices in $\mathcal{N}^{\downarrow}$}
	\State $\mathcal{B}_1, \mathcal{B}_2, \dots, \mathcal{B}_B \gets \text{split}(\mathcal{N}^{\downarrow})$ \Comment{Split $\mathcal{N}^{\downarrow}$ into $B=\lceil N/K \rceil$ batches}
	\State labels[$\mathcal{B}_1$] $\gets [1 \dots K]$ \Comment{Assign one object from $\mathcal{B}_1$ to each anticluster}

	\For{$k = 1$ \textbf{to} $K$} \Comment{Initialize the centroids of the $K$ anticlusters}
		\State $i \gets \mathcal{B}_1[k]$
		\State $\bm{\mu}_k \gets \bm{x}_i$ 
	\EndFor
	
	\For{$b = 2$ \textbf{to} $B$}
		\State $\text{costmatrix} \gets \text{empty matrix of size } |\mathcal{B}_b| \times K$ \Comment{Initialize cost matrix}	
		\For{$j=1$ \textbf{to} $|\mathcal{B}_b|$} \Comment{Compute squared Euclidean distances}
			\For{$k=1$ \textbf{to} $K$}
				\State $\text{costmatrix}[j][k] \gets \|\bm{x}_{\mathcal{B}_b[j]} - \bm{\mu}_k\|_2^2$ 
			\EndFor
		\EndFor
		\State $\text{new\_labels} \gets \text{SOLVE\_ASSIGNMENT\_PROBLEM}(\text{costmatrix})$
		\For{$j=1$ \textbf{to} $|\mathcal{B}_b|$}
			\State $k \gets \text{new\_labels}[j]$
			\State $i \gets \mathcal{B}_b[j]$
			\State $\bm{\mu}_k \gets \text{UPDATE\_CENTROID}(b, \bm{\mu}_k, \bm{x}_i)$ 
		\EndFor
		
		\State $\text{labels}[\mathcal{B}_b] \gets \text{new\_labels}$		
	\EndFor	

    \State \textbf{return} labels
\EndFunction
\Statex
\end{algorithmic} \label{alg:base}

\begin{algorithmic}
\Function{solve\_assignment\_problem}{\text{costmatrix}}
	\State assignment $\gets$ \text{LAPJV}(\text{costmatrix}) \Comment{LAPJV is a variant of the Jonker-Volgenant algorithm}
	\State new\_labels $\gets$ empty vector
    \For{each (row\_id, col\_id) in assignment}
        \State new\_labels[row\_id] $\gets$ col\_id
    \EndFor	
    \State \Return new\_labels
\EndFunction
\end{algorithmic}
\vspace{0.5cm}
\begin{algorithmic}
\Function{update\_centroid}{\text{counter}, $\bm{\mu}_k$, $\bm{x}_i$}
    \State \Return $\bm{\mu}_k + \frac{1}{\text{counter}}(\bm{x}_i - \bm{\mu}_k)$
\EndFunction
\end{algorithmic}
\end{algorithm}

\subsection{The \ABA~Algorithm With Categories}\label{sec:algorithm:categories}
\begin{figure}
	\centering\includegraphics[width=\textwidth]{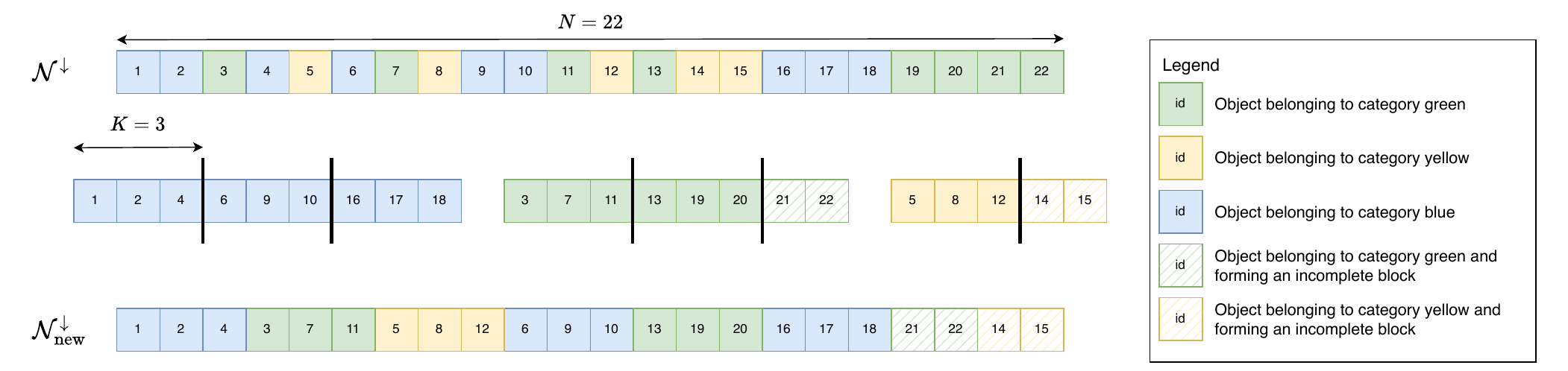}
	\caption{Illustration of how to rearrange the sorted list $\mathcal{N}^{\downarrow}$ to evenly distribute the categories of a categorical feature across anticlusters based on an example with $N=22$ objects and $K=3$ anticlusters.\label{fig:list_adjustment_aba_with_categories}}
\end{figure}

We also propose a variant of the \ABA~algorithm for the anticlustering problem with categories (see~Section~\ref{sec:problem}). The base variant of the \ABA~algorithm requires two modifications to account for the additional constraint that all objects of the same category need to be evenly distributed across anticlusters. First, the objects in the sorted list $\mathcal{N}^{\downarrow}$ need to be rearranged as follows. The list is divided into $G$ sublists $\mathcal{N}^{\downarrow}_g$ ($g \in G$), where each $\mathcal{N}^{\downarrow}_g$ contains all objects belonging to category $g$, preserving their original order. Each $\mathcal{N}^{\downarrow}_g$ is then partitioned into consecutive blocks of size $K$. If the number of objects in $\mathcal{N}^{\downarrow}_g$ is not a multiple of $K$, the final block contains fewer than $K$ objects. The rearranged list is obtained by sequentially concatenating all full ($K$-sized) blocks from the $G$ lists in an alternating fashion, always selecting the next available block according to the sorting. The remaining (incomplete) blocks are appended at the end in the same alternating order. Figure~\ref{fig:list_adjustment_aba_with_categories} illustrates for an example with $N=22$ objects and $K=3$ anticlusters how to construct the rearranged list $\mathcal{N}^{\downarrow}_\text{new}$. Second, when splitting the rearranged list $\mathcal{N}^{\downarrow}_\text{new}$ into batches $\mathcal{B}_1$ to $\mathcal{B}_B$, the last batches may contain objects that belong to different categories. Because of this, we need to keep track of the number of objects of category $g \in G$ assigned to each anticluster $k \in K$ during the iterations. When assigning object $i$ to anticluster $k$ would violate the corresponding upper bound $\left\lceil \frac{|\mathcal{N}_g|}{K} \right\rceil$, the cost matrix entry that corresponds to the distance between $\bm{x}_i$ and $\bm{\mu}_k$ must be set to a sufficiently large negative value, such as the negative of the maximum entry in the cost matrix, before solving the assignment problem to ensure that the upper bound is never exceeded. 

\subsection{Computational Speed-Up with Hierarchical Decomposition}\label{sec:algorithm:hierarchical}
\begin{figure}
	\centering\includegraphics[width=0.7\textwidth]{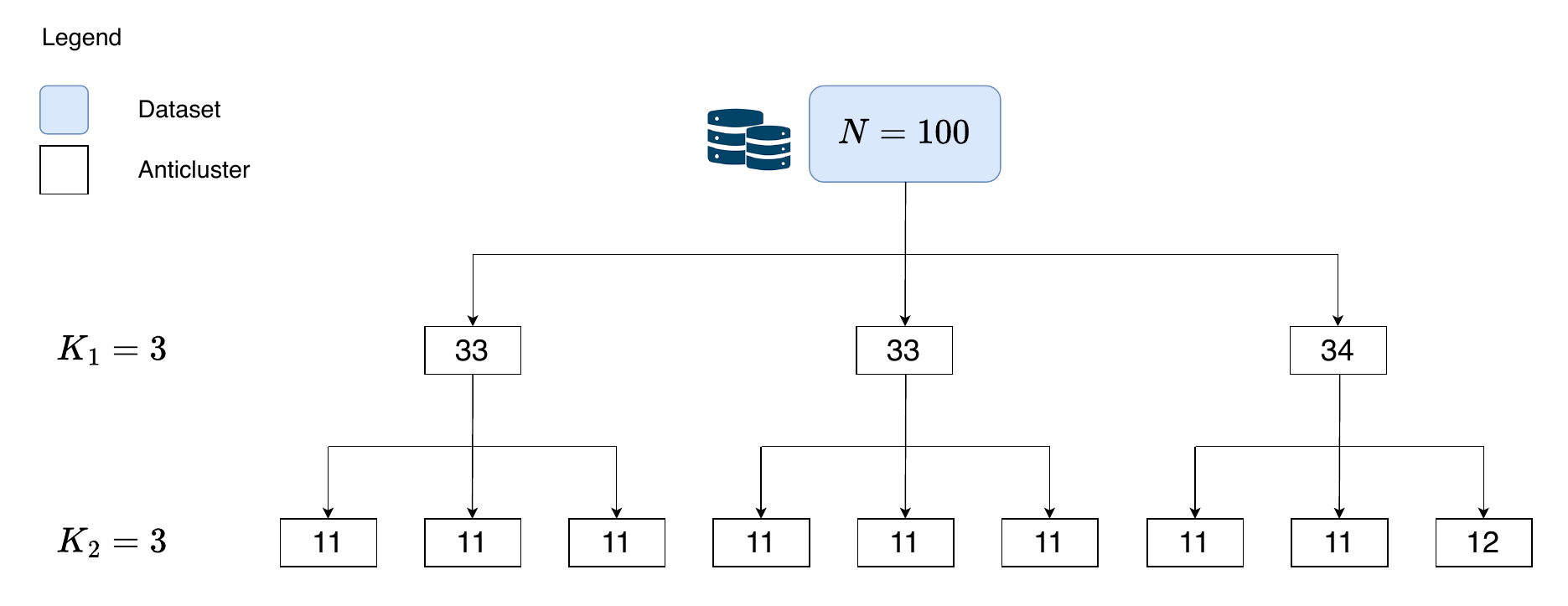}
	\caption{Illustration of the hierarchical decomposition strategy for a dataset with $N=100$ objects and $K=9$ anticlusters.\label{fig:computational_speedup}}
\end{figure}

The size of the assignment problems solved during the execution of the \ABA~algorithm depends on~$K$. For large-scale instances with large values of $K$, solving the assignment problems can become computationally prohibitive. For such instances, we propose a hierarchical decomposition strategy that achieves substantial reductions in running time while maintaining high solution quality. The idea is to construct the anticlusters in a hierarchical manner, i.e., first create $K_1$ anticlusters and then subdivide each of the $K_1$ anticlusters again into $K_2$ anticlusters. This will yield $K_1 \times K_2$ anticlusters. The number of levels $L$ and the values $K_1, K_2, \ldots, K_L$ can be specified by the user. As stated in Lemma \ref{lem:balanced-parameters}, a balanced decomposition in which the values of $K_{\ell}$ are as similar as possible,
will deliver the lowest complexity for $L$ levels. 
We also observe experimentally, in Figure~\ref{fig:barplot},  that such a choice does not noticeably diminish solution quality. Note that the subproblems that result from the decomposition can be solved in parallel, leading to further improvements in the computational speed of the algorithm on multi-core machines. Figure~\ref{fig:computational_speedup} illustrates the hierarchical decomposition strategy based on an example with $N=100$ objects and $K=9$ anticlusters. The hierarchical decomposition strategy proposed above still guarantees that the sizes of the final anticlusters lie within the prescribed bounds $\floor{N/K}$ and $\ceil{N/K}$. This follows because each split is balanced, meaning every resulting anticluster receives either the baseline number of objects or one additional object, and these additional objects are simply passed down the recursion tree. Since no split can introduce an imbalance larger than one object, the final anticluster sizes also differ by at most one object. 

\subsection{The Complexity of the \ABA~Algorithm and the Hierarchical Decomposition}\label{sec:algorithm:complexity}

The time complexity of the base variant of the \ABA{} algorithm is determined by three main steps. First, computing the global centroid of the objects and their distances to this centroid requires $O(ND)$ operations. Second, sorting the objects in decreasing order of their distance to the global centroid requires $O(N \log N)$ time. Third, the algorithm solves $\ceil{N/K}-1$ assignment problems using an implementation of the algorithm of \cite{jonker1987shortest} called \textit{LAPJV}. Since each assignment problem requires $O(K^3)$ operations, this step requires a total of $O(\frac{N}{K}K^3) = O(NK^2)$ time. Hence, the overall time complexity of the \ABA{} algorithm is $O(ND) + O(N \log N) + O(NK^2) 
\text{ or } O\bigl(N(D + \log N + K^2)\bigr)$. Since the calls to LAPJV dominate the running time, the running time is $O(NK^2)$. The time complexity is the same for the \ABA~algorithm with categories as the modifications with respect to the base variant do not change the overall time complexity. 

We next discuss the complexity of the hierarchical decomposition. To illustrate the total complexity expression consider the example of a hierarchical decomposition with $L=3$ levels and the choice of parameters $K_1,K_2,K_3$. The complexity of \ABA\ at each level is as follows:

\begin{itemize}
\item Level 1: \ABA\ is applied once to the full instance with parameter $K_1$, with complexity, $T_1 = O(NK_1^2)$, producing $K_1$ subproblems.
\item Level 2: Each of the $K_1$ subproblems here contains approximately $N/K_1$ objects and is solved with parameter $K_2$, partitioning each one of the $K_1$ subproblems to $K_2$ subproblems, for a total of $K_1K_2$ subproblems. The total cost at this level is then
$T_2 = K_1 \cdot O\!\left(\frac{N}{K_1}K_2^2\right) = O(NK_2^2).$
\item Level 3: Each of the $K_1K_2$ resulting subproblems contains approximately $N/(K_1K_2)$ objects and is solved with parameter $K_3$.
The total cost at this level is, $T_3 = K_1K_2 \cdot O\!\left(\frac{N}{K_1K_2}K_3^2\right) = O(NK_3^2).$
\end{itemize}

Summing over all three levels yields $T_{\mathrm{total}} = O\bigl(N(K_1^2 + K_2^2 + K_3^2)\bigr).$ Extrapolating from this $3$-level example, the complexity of the hierarchical decomposition with $L$ levels and parameters $K_1,\ldots ,K_L$ is $T_\ell= O(NK_\ell^2) $ for level $\ell =1,\ldots, L$. The total complexity is then $\sum_{\ell=1}^{L} T_\ell = O\!\left(N\sum_{\ell=1}^{L} K_\ell^2\right).$ The lemma below states that the selection of equal parameter values for $K_{\ell}$ minimizes the total complexity expression.

The proof of the following lemma is straightforward and therefore omitted. 

\begin{lemma}\label{lem:balanced-parameters}
    The total complexity of the hierarchical decomposition with $L$ levels $K_1$ through $K_L$, with $K_\ell$ denoting the number of anticlusters in level $\ell$, $\sum_{\ell=1}^{L} T_\ell = O\!\left(N\sum_{\ell=1}^{L} K_\ell^2\right)$, is minimized for $K_1=\ldots =K_L= K^{1/L}$. With this optimal choice, the total complexity of the hierarchical decomposition is
$ O\!\left(N \, L \, K^{2/L}\right)$.
\end{lemma}

\section{Experimental Results and Analysis}\label{sec:experiment}
In this section, we compare the performance of \ABA~with state-of-the-art algorithms in three experiments. First, we evaluate \ABA~against the state-of-the-art algorithm of \cite{papenberg2021using} for large-scale anticlustering in Euclidean space. Second, we assess the effectiveness of the hierarchical decomposition strategy. Third, we compare \ABA~to \METIS\ (see \citealt{karypis1998fast}), the leading software for the balanced $K$-cut problem. The section is organized as follows. Section~\ref{experiment:data} introduces the datasets, Section~\ref{experiment:algorithms} presents the benchmark algorithms, and Sections~\ref{experiment:results_standard}, \ref{experiment:hierarchical_decomposition}, and \ref{experiment:balanced_cut} report the experimental results. The results of an additional experiment on Euclidean anticlustering with categories, reported in the Electronic Companion on \href{https://github.com/phil85/aba-results}{\color{blue}GitHub}, show that \ABA~outperforms the algorithm of \cite{croella2025strong} in both solution quality and running time, and matches the solution quality of the  algorithms from \cite{papenberg2021using} while being substantially faster. We implemented the \ABA~algorithm in C++ and used the \href{https://github.com/src-d/lapjv}{\color{blue}LAPJV} algorithm for solving the assignment problems. Our C++ source code is available on \href{https://github.com/j-yang-932/aba_Cpp}{\color{blue}GitHub}. We also provide a Python implementation of the \ABA~algorithm on \href{https://github.com/phil85/aba}{\color{blue}GitHub}. All computations were executed on an HP workstation with two Intel Xeon CPUs with clock speed 3.30 GHz and 256 GB of RAM. 


\subsection{Datasets}\label{experiment:data} 
We use a variety of datasets that differ in the number of objects ($N$), the number of features ($D$), and the data type (tabular or image). Table~\ref{tbl:datasets} lists each dataset along with its source. Besides datasets from prior anticlustering research, we include additional datasets commonly used in machine learning. We applied the following preprocessing steps. For datasets with categorical features, we encoded the categorical features as a set of binary features using a one-hot encoding scheme, creating one binary feature for each category. For datasets that included nonbinary features, we standardized all features by subtracting the mean and dividing by the standard deviation. Additional preprocessing was performed for specific datasets as described next. For the \texttt{Travel} dataset, we drop the \textit{User} feature and remove objects with missing values. For the \texttt{Npi} dataset, we preprocessed the data in the same way as in \cite{papenberg2021using}, i.e., we removed objects with missing values and kept only the binary features representing the answers of the participants. For the \texttt{Creditcard} dataset, we leave out the \textit{ID} feature. For the \texttt{Census} dataset, we drop the \textit{caseid} feature. For the \texttt{Plants} dataset, we converted the original data into a binary matrix, with rows representing plants and columns representing states. An entry of one indicates that the plant occurs in the corresponding state. In the image datasets \texttt{Cifar10}, \texttt{Mnist}, \texttt{Imagenet8}, and \texttt{Imagenet32}, the features represent pixel intensities with one (grayscale) or three (RGB) values ranging from 0 to 255. As is commonly done with image data, we scaled the features by dividing them by 255 instead of standardizing them, bringing all values into the range [0,1]. Note that the \texttt{Imagenet8} and \texttt{Imagenet32} datasets are downsampled versions of the original ImageNet dataset, with image resolutions of 8$\times$8 and 32$\times$32 pixels, respectively. For the \texttt{Finance} dataset, we eliminated the features \textit{step}, \textit{nameOrig}, and \textit{nameDest}.

\begin{table}
	\footnotesize
    \centering
	\setlength{\tabcolsep}{10pt}
	\begin{tabular}{lrrll}
\toprule
Dataset & Objects ($N$) & Features ($D$) & Type & Source \\
\midrule
Travel & 5,454 & 24 & Tabular & \href{https://archive.ics.uci.edu/dataset/485/tarvel+review+ratings}{\color{blue}UCI ML Repository} \\
Npi & 10,440 & 40 & Tabular & \href{https://openpsychometrics.org/_rawdata/}{\color{blue}Online personality tests} \\
Creditcard & 30,000 & 24 & Tabular & \href{https://archive.ics.uci.edu/dataset/350/default+of+credit+card+clients}{\color{blue}UCI ML Repository} \\
Adult & 32,561 & 110 & Tabular & \href{https://archive.ics.uci.edu/dataset/2/adult}{\color{blue}UCI ML Repository} \\
Plants & 34,781 & 70 & Tabular & \href{https://archive.ics.uci.edu/dataset/180/plants}{\color{blue}UCI ML Repository} \\
Bank & 45,211 & 53 & Tabular & \href{https://archive.ics.uci.edu/dataset/222/bank+marketing}{\color{blue}UCI ML Repository} \\
Cifar10 & 50,000 & 3,072 & Image & \href{https://www.cs.toronto.edu/~kriz/cifar.html}{\color{blue}UCI ML Repository} \\
Mnist & 60,000 & 784 & Image & \href{https://www.kaggle.com/datasets/hojjatk/mnist-dataset?select=t10k-labels.idx1-ubyte}{\color{blue}Kaggle} \\
Survival & 110,204 & 4 & Tabular & \href{https://archive.ics.uci.edu/dataset/827/sepsis+survival+minimal+clinical+records}{\color{blue}UCI ML Repository} \\
Diabetes & 253,680 & 22 & Tabular & \href{https://archive.ics.uci.edu/dataset/891/cdc+diabetes+health+indicators}{\color{blue}UCI ML Repository} \\
Music & 515,345 & 91 & Tabular & \href{https://archive.ics.uci.edu/dataset/203/yearpredictionmsd}{\color{blue}Kaggle} \\
Covtype & 581,012 & 55 & Tabular & \href{https://archive.ics.uci.edu/dataset/31/covertype}{\color{blue}UCI ML Repository} \\
Imagenet8 & 1,281,167 & 192 & Image & \href{https://www.image-net.org/index.php}{\color{blue}ImageNet Image Database} \\
Imagenet32 & 1,281,167 & 3,072 & Image & \href{https://www.image-net.org/index.php}{\color{blue}ImageNet Image Database} \\
Census & 2,458,285 & 68 & Tabular & \href{https://archive.ics.uci.edu/dataset/116/us+census+data+1990}{\color{blue}UCI ML Repository} \\
Finance & 6,362,620 & 12 & Tabular & \href{https://www.kaggle.com/datasets/sriharshaeedala/financial-fraud-detection-dataset}{\color{blue}Kaggle} \\
\bottomrule
\end{tabular}

	\caption{Overview of datasets for Euclidean anticlustering\label{tbl:datasets} }
\end{table}

For the additional experiment on Euclidean anticlustering with categories, reported in the Electronic Companion on \href{https://github.com/phil85/aba-results}{\color{blue}GitHub}, we use the datasets that were considered in \cite{croella2025strong}. These datasets are listed in Table~\ref{tbl:datasets_categories}. The same preprocessing steps as described above were applied to these datasets. 

\begin{table}
	\footnotesize
    \centering
	\setlength{\tabcolsep}{10pt}
	\begin{tabular}{lrrrll}
\toprule
Dataset & Objects ($N$) & Features ($D$) & Number of categories & Type & Source \\
\midrule
Abalone & 4,177 & 10 & 3 & Tabular & \href{https://archive.ics.uci.edu/dataset/1/abalone}{\color{blue}UCI ML Repository} \\
Facebook & 7,050 & 13 & 3 & Tabular & \href{https://archive.ics.uci.edu/dataset/488/facebook+live+sellers+in+thailand}{\color{blue}UCI ML Repository} \\
Frogs & 7,195 & 22 & 4 & Tabular & \href{https://archive.ics.uci.edu/dataset/406/anuran+calls+mfccs}{\color{blue}UCI ML Repository} \\
Electric & 10,000 & 12 & 3 & Tabular & \href{https://archive.ics.uci.edu/dataset/471/electrical+grid+stability+simulated+data}{\color{blue}UCI ML Repository} \\
Pulsar & 17,898 & 8 & 2 & Tabular & \href{https://archive.ics.uci.edu/dataset/372/htru2}{\color{blue}UCI ML Repository} \\
\bottomrule
\end{tabular}

	\caption{Overview of datasets for Euclidean anticlustering with categories for which \cite{croella2025strong} added categories \label{tbl:datasets_categories}}
\end{table}

\subsection{Benchmark Algorithms}\label{experiment:algorithms}
The benchmark algorithms used in our experiments include the \texttt{fast-anticlustering} algorithm from the R package of \cite{papenberg2021using}, the default balanced $K$-cut algorithm from the \METIS\ software package (see \citealt{karypis1998fast}), and a random partitioning procedure. In the additional experiment on Euclidean anticlustering with categories reported on \href{https://github.com/phil85/aba-results}{\color{blue}GitHub}, we also use the integer linear programming model of \cite{croella2025strong} as a benchmark algorithm. The R package \texttt{anticlust} of \cite{papenberg2021using} includes exact integer programming algorithms or exchange heuristics. 
The exact algorithms are only able to solve small problem instances with up to 70 objects in acceptable running time and therefore are not tested here. 
From the two exchange-based heuristics available in the \texttt{anticlust} package, we use the heuristic called \texttt{fast-anticlustering} as this is the most scalable one. 
In \cite{papenberg2021using}, the \texttt{fast-anticlustering} algorithm is applied with 5 exchange partners determined by nearest neighbor search. We run it with the same settings and in addition with 5, 50, and 500 randomly generated exchange partners. To the best of our knowledge, the \texttt{fast-anticlustering} algorithm is the leading algorithm for large-scale anticlustering. We applied the balanced $K$-cut algorithm from the \METIS\ software package with its default control parameter values. It requires an input file that specifies all neighbors for each node of the graph together with the corresponding edge weights as integers. The tests of \METIS\ versus \ABA~are described in Section~\ref{experiment:balanced_cut}. Table~\ref{tbl:algorithms} lists each algorithm together with its abbreviation, full name, source publication, programming language, and a link to its source code.

\begin{table}
	\footnotesize
	\setlength{\tabcolsep}{8pt}
	\begin{tabular}{lllll} \toprule
		Abbr.\  & Name                                        & Paper                              & Code & Link                                                       \\ \midrule
		\ABA\    & Assignment-based anticlustering             & This paper                         & C++    & \href{https://github.com/j-yang-932/aba_Cpp}{\color{blue}GitHub}                                        \\
		P-N5      & Fast-anticlustering with 5 nearest-neighbor exchange partners   & \cite{papenberg2021using} & C    & \href{https://github.com/m-Py/anticlust/tree/main}{\color{blue}GitHub} \\		
		P-R5      & Fast-anticlustering with 5 random exchange partners   & \cite{papenberg2021using} & C    & \href{https://github.com/m-Py/anticlust/tree/main}{\color{blue}GitHub} \\
		P-R50     & Fast-anticlustering with 50 random exchange partners  & \cite{papenberg2021using} & C    & \href{https://github.com/m-Py/anticlust/tree/main}{\color{blue}GitHub} \\
		P-R500     & Fast-anticlustering with 500 random exchange partners & \cite{papenberg2021using} & C    & \href{https://github.com/m-Py/anticlust/tree/main}{\color{blue}GitHub} \\
		MILP    & MILP model of AVOC algorithm                & \cite{croella2025strong}           & C    & \href{https://github.com/AnnaLivia/AVOC}{\color{blue}GitHub}           \\
		\METIS\   & Balanced $K$-cut algorithm from \METIS\       & \cite{karypis1998fast}             & C    & \href{https://github.com/KarypisLab/METIS}{\color{blue}GitHub}         \\
		Rand    & Random partitioning                         & This paper                                & C++    & \href{https://github.com/j-yang-932/aba_Cpp}{\color{blue}GitHub}                                                        \\
		\bottomrule
	\end{tabular}
	\caption{Overview of all algorithms used in the experimental analysis\label{tbl:algorithms}}
\end{table}

\subsection{Comparison to Leading Algorithms on Standard Anticlustering Problems}\label{experiment:results_standard}
In this experiment, we compare the \ABA~algorithm with the \texttt{fast-anticlustering} algorithm of \cite{papenberg2021using} using the 16 datasets from Table~\ref{tbl:datasets}. For each of the 16 datasets, we generated multiple problem instances by varying the number of anticlusters $K$. For the smallest dataset named Travel with less than $N=10{,}000$ objects, we created six problem instances by setting $K = \{2, 5, 50, 500, 1{,}000, 2{,}000\}$, for the other datasets with $N=10{,}000$ or more objects, we created seven problem instances by setting $K = \{2, 5, 50, 500, 1{,}000, 2{,}000, 5{,}000\}$. This yields a total of 111 problem instances. We applied \texttt{fast-anticlustering} with the same configuration as in \cite{papenberg2021using}, using 5 exchange partners determined via nearest-neighbor search (denoted P-N5). In addition, we applied \texttt{fast-anticlustering} with 5, 50, and 500 randomly chosen exchange partners (P-R5, P-R50, P-R500). As a baseline, we use random partitioning (Rand), which randomly permutes the objects and assigns them cyclically to $K$ anticlusters. We use the objective function values and the running time as performance metrics. The \ABA~algorithm is deterministic, whereas the other algorithms are randomized and require a random seed. We therefore ran each benchmark algorithm three times per problem instance and report the average performance metrics. 
We stopped a run if a time limit of two hours has been reached. For large instances, we applied \ABA~with hierarchical decomposition. Table~\ref{tbl:hierarchical} specifies the settings we used for the hierarchical decomposition as a function of the number of objects ($N$) and the number of anticlusters ($K$). The detailed results for all 111 problem instances are available on \href{https://github.com/phil85/aba-results}{\color{blue}GitHub}. Due to limited space, we provide here a summary of the findings. First, we present the results for the Npi dataset with all values of $K=\{2, 5, 50, 500, 1{,}000, 2{,}000, 5{,}000\}$. This dataset was also used in \cite{papenberg2021using}. Then, we present results for all datasets with $K=5$ and $K=1{,}000$. 
\begin{table}
	\scriptsize
	\centering
	\begin{tabular}{lrrrrrrr}
		& \multicolumn{7}{c}{$K$} \\ \cmidrule(lr){2-8}
		& 2 & 5 & 50 & 500 & $1{,}000$ & $2{,}000$ & $5{,}000$ \\ \midrule
		$N\leq 50{,}000$ & --- & --- & --- & --- & (2$\times$500) & (4$\times$500) & (10$\times$500) \\
		$N > 50{,}000$ & --- & --- & --- & (4$\times$125) & (8$\times$125) & (16$\times$125) & (40$\times$125) \\
	\end{tabular}
	\caption{The settings used for hierarchical decomposition as a function of the number of objects ($N$) and the number of anticlusters ($K$). A dash (---) indicates that hierarchical decomposition was not applied. The expression $(K_1 \times K_2)$ denotes a decomposition with two levels and $K=K_1 \times K_2$.\label{tbl:hierarchical}}
\end{table}

Table~\ref{tbl:npi} presents the results for the Npi dataset. The first column lists the number of anticlusters~$K$. The second column reports the objective function values for the solutions obtained by the \ABA~algorithm (ofv \ABA), computed here as the sum of squared Euclidean distances between the objects and their corresponding anticluster centroid. Columns~3--7 show the percentage deviation of the objective function values obtained by the benchmark algorithms from that of \ABA. A negative deviation indicates that the \ABA~algorithm outperformed the respective algorithm. Column~8 gives the running time of the \ABA~algorithm in seconds, while columns 9--12 report the percentage deviations of the running time of the \texttt{fast-anticlustering} algorithms from that of \ABA. A positive deviation indicates that the respective algorithm is slower than the \ABA~algorithm. The bottom row of the table reports the average values. For each instance (row), the best solution quality and the best running time are highlighted in bold. The main conclusions from Table~\ref{tbl:npi} are that the \ABA~algorithm consistently provides the best solution quality and is substantially faster than the \texttt{fast-anticlustering} algorithms. For small values of $K$, all tested algorithms deliver similar solution quality. For larger values of $K$, however, the \ABA~algorithm provides substantial improvements in terms of solution quality. In particular, for $K=5{,}000$, the \ABA~algorithm provides a solution that is, on average, more than 16\% better than the solutions produced by P-N5 and more than 30\% better than the solutions produced by random partitioning.   

\begin{table}
	\scriptsize
	\setlength{\tabcolsep}{4.5pt}
	\begin{tabular}{rrrrrrrrrrrr}
\toprule 
 \multicolumn{12}{l}{Dataset: Npi ($N$=10,440, $D$=40)} \\ \midrule
\multicolumn{2}{c}{} & \multicolumn{5}{c}{Deviation from ofv \textsf{ABA} [\%]} &  & \multicolumn{4}{c}{Deviation from cpu \textsf{ABA} [\%]} \\ \cmidrule(lr){3-7} \cmidrule(lr){9-12}
$K$ & ofv \textsf{ABA} & P-N5 & P-R5 & P-R50 & P-R500 & Rand & cpu \textsf{ABA} [s] & P-N5 & P-R5 & P-R50 & P-R500 \\
\cmidrule(lr){1-1} \cmidrule(lr){2-7} \cmidrule(lr){8-12}
2 & \bfseries 87,153.6218 & \bfseries 0.0000 & \bfseries 0.0000 & \bfseries 0.0000 & \bfseries 0.0000 & -0.0123 & \bfseries 0.004 & 241,508.54 & 31,899.67 & 42,924.13 & 120,903.36 \\
5 & \bfseries 87,153.5939 & \bfseries 0.0000 & \bfseries 0.0000 & \bfseries 0.0000 & \bfseries 0.0000 & -0.0320 & \bfseries 0.004 & 255,033.12 & 33,576.33 & 47,813.83 & 156,944.45 \\
50 & \bfseries 87,151.7909 & -0.0030 & -0.0104 & -0.0032 & -0.0008 & -0.4710 & \bfseries 0.017 & 56,553.53 & 7,444.68 & 10,988.05 & 39,133.62 \\
500 & \bfseries 86,977.7929 & -1.0587 & -1.1193 & -0.5510 & -0.2977 & -4.5654 & \bfseries 0.200 & 4,636.56 & 530.12 & 833.03 & 3,253.13 \\
1000 & \bfseries 86,399.2545 & -3.0555 & -2.9480 & -1.6556 & -0.9476 & -8.7383 & \bfseries 0.198 & 4,705.28 & 544.11 & 843.40 & 3,285.61 \\
2000 & \bfseries 84,068.5000 & -7.3996 & -6.6095 & -4.0046 & -2.3669 & -16.0867 & \bfseries 0.162 & 5,755.62 & 689.38 & 1,055.61 & 4,070.11 \\
5000 & \bfseries 67,221.8333 & -16.7954 & -13.5452 & -7.2405 & -3.0945 & -32.5496 & \bfseries 0.121 & 7,786.34 & 956.07 & 1,350.11 & 4,402.66 \\\cmidrule(lr){1-1} \cmidrule(lr){2-7} \cmidrule(lr){8-12} 
 Average &  & -4.0446 & -3.4618 & -1.9221 & -0.9582 & -8.9222 &  & 82,282.71 & 10,805.76 & 15,115.45 & 47,427.56 \\
\bottomrule \\ [-1.5ex] \multicolumn{7}{l}{\textsf{ABA} is the contribution in this paper}
\end{tabular}

	\caption{Comparing solution quality and running times for instances of the Npi dataset of \ABA, P-N5, P-R5, P-R50, P-R500, and random. Reported are the objective function values (ofv) for the \ABA~algorithm and the percentage deviations from these values for the other algorithms (positive deviation means improving on \ABA). Also reported are the running times of the \ABA~algorithm in seconds (cpu) and the percentage deviations from these values for the other algorithms (negative deviation means improving on \ABA). \label{tbl:npi}}
\end{table}

Table~\ref{tbl:aggregated_results_k5_ofv_cpu} presents the results for all datasets when $K=5$. The structure of the table is similar to that of Table~\ref{tbl:npi} only that the first three columns list the name of the dataset, the number of objects ($N$), and the number of features ($D$). A dash (---) means that the respective algorithm did not return a solution within the prescribed time limit of two hours in any of the three runs. To compute the average, the bottom row of the table, we exclude the dash-entries and take the average only across the entries where the respective algorithm computed a solution. From Table~\ref{tbl:aggregated_results_k5_ofv_cpu}, we observe that the \texttt{fast-anticlustering} algorithms (P-N5, P-R5, P-R50, P-R500) do not find solutions for all instances within the two hours time limit. In contrast, the \ABA~algorithm produces solutions to all instances in less than a second or within a few seconds. Since $K$ is small, all algorithms achieve comparable solution quality, although Rand performs slightly worse. However, \ABA~is substantially faster. While it produces the solution for each dataset in under a second or just a few seconds, P-N5, P-R5, P-R50, and P-R500 require considerably longer running times, with average running times up to three orders of magnitude longer than those of \ABA. 

\begin{table}
	\tiny
	\setlength{\tabcolsep}{4.5pt}
	\begin{tabular}{lrrrrrrrrrrrrr}
\toprule 
 \multicolumn{14}{l}{$K$= 5} \\ \midrule
\multicolumn{4}{c}{} & \multicolumn{5}{c}{Deviation from ofv \textsf{ABA} [\%]} &  & \multicolumn{4}{c}{Deviation from cpu \textsf{ABA} [\%]} \\ \cmidrule(lr){5-9} \cmidrule(lr){11-14}
Dataset & $N$ & $D$ & ofv \textsf{ABA} & P-N5 & P-R5 & P-R50 & P-R500 & Rand & cpu \textsf{ABA} [s] & P-N5 & P-R5 & P-R50 & P-R500 \\
\cmidrule(lr){1-3} \cmidrule(lr){4-9} \cmidrule(lr){10-14}
Travel & 5,454 & 24 & 130,871.89 & \bfseries 0.0001 & 0.0000 & \bfseries 0.0001 & \bfseries 0.0001 & -0.0192 & \bfseries 0.002 & 45,104.8 & 34,479.4 & 44,869.6 & 129,328.6 \\
Npi & 10,440 & 40 & \bfseries 87,153.59 & \bfseries 0.0000 & \bfseries 0.0000 & \bfseries 0.0000 & \bfseries 0.0000 & -0.0320 & \bfseries 0.004 & 255,033.1 & 33,576.3 & 47,813.8 & 156,944.4 \\
Creditcard & 30,000 & 24 & \bfseries 719,975.99 & \bfseries 0.0000 & \bfseries 0.0000 & \bfseries 0.0000 & \bfseries 0.0000 & -0.0160 & \bfseries 0.010 & 76,110.7 & 36,331.0 & 47,208.7 & 119,590.6 \\
Adult & 32,561 & 110 & \bfseries 3,581,588.43 & -0.0048 & \bfseries 0.0000 & \bfseries 0.0000 & \bfseries 0.0000 & -0.0121 & \bfseries 0.025 & 48,457.0 & 18,036.4 & 32,013.6 & 143,843.5 \\
Plants & 34,781 & 70 & \bfseries 258,111.51 & \bfseries 0.0000 & \bfseries 0.0000 & \bfseries 0.0000 & \bfseries 0.0000 & -0.0083 & \bfseries 0.019 & 94,101.5 & 25,195.4 & 37,595.0 & 125,792.1 \\
Bank & 45,211 & 53 & \bfseries 2,396,129.86 & -0.0017 & \bfseries 0.0000 & \bfseries 0.0000 & \bfseries 0.0000 & -0.0051 & \bfseries 0.024 & 65,465.5 & 26,216.3 & 39,325.8 & 126,328.3 \\
Cifar10 & 50,000 & 3,072 & \bfseries 9,520,045.51 & \bfseries 0.0000 & -0.0001 & \bfseries 0.0000 & \bfseries 0.0000 & -0.0105 & \bfseries 0.682 & 610,874.4 & 10,035.1 & 79,509.9 & 761,007.1 \\
Mnist & 60,000 & 784 & \bfseries 3,163,501.01 & \bfseries 0.0000 & \bfseries 0.0000 & \bfseries 0.0000 & \bfseries 0.0000 & -0.0062 & \bfseries 0.223 & 1,362,505.0 & 9,737.3 & 57,277.3 & 528,621.2 \\
Survival & 110,204 & 4 & \bfseries 440,812.00 & -0.0038 & \bfseries 0.0000 & \bfseries 0.0000 & \bfseries 0.0000 & -0.0026 & \bfseries 0.024 & 60,549.7 & 58,127.1 & 72,821.1 & 123,547.4 \\
Diabetes & 253,680 & 22 & \bfseries 5,580,938.00 & -0.0003 & \bfseries 0.0000 & \bfseries 0.0000 & \bfseries 0.0000 & -0.0015 & \bfseries 0.145 & 87,496.1 & 24,957.1 & 32,754.8 & 76,951.9 \\
Music & 515,345 & 91 & \bfseries 46,896,303.99 & \NA & \bfseries 0.0000 & \bfseries 0.0000 & \bfseries 0.0000 & -0.0006 & \bfseries 0.483 & \NA & 16,723.4 & 30,304.8 & 155,567.6 \\
Covtype & 581,012 & 55 & \bfseries 31,955,602.87 & -0.0005 & \bfseries 0.0000 & \bfseries 0.0000 & \bfseries 0.0000 & -0.0002 & \bfseries 0.394 & 28,849.3 & 22,816.1 & 36,154.8 & 142,034.4 \\
Imagenet8 & 1,281,167 & 192 & \bfseries 13,327,181.18 & \NA & \bfseries 0.0000 & \bfseries 0.0000 & \bfseries 0.0000 & -0.0003 & \bfseries 1.872 & \NA & 12,095.5 & 30,794.8 & 203,514.9 \\
Imagenet32 & 1,281,167 & 3,072 & \bfseries 265,674,709.96 & \NA & \NA & \NA & \NA & -0.0003 & \bfseries 17.733 & \NA & \NA & \NA & \NA \\
Census & 2,458,285 & 68 & \bfseries 167,163,312.00 & \NA & \bfseries 0.0000 & \bfseries 0.0000 & \bfseries 0.0000 & -0.0002 & \bfseries 2.342 & \NA & 16,299.1 & 28,130.4 & 128,512.1 \\
Finance & 6,362,620 & 12 & \bfseries 76,351,427.75 & \bfseries 0.0000 & \bfseries 0.0000 & \bfseries 0.0000 & \NA & \bfseries 0.0000 & \bfseries 4.532 & 24,387.9 & 20,434.0 & 26,493.8 & \NA \\\cmidrule(lr){1-3} \cmidrule(lr){4-9} \cmidrule(lr){10-14} 
 Average & & &  & -0.0009 & 0.0000 & 0.0000 & 0.0000 & -0.0072 &  & 229,911.3 & 24,337.3 & 42,871.2 & 208,684.6 \\
\bottomrule \\ [-1.5ex] \multicolumn{7}{l}{\textsf{ABA} is the contribution in this paper}
\end{tabular}

	\caption{Comparing solution quality and running times for $K=5$ of \ABA, P-N5, P-R5, P-R50, P-R500, and random. Reported are the objective function values (ofv) for the \ABA~algorithm and the percentage deviations from these values for the other algorithms (positive deviation means improving on \ABA). Also reported are the running times of the \ABA~algorithm in seconds (cpu) and the percentage deviations from these values for the other algorithms (negative deviation means improving on \ABA). A dash indicates that the algorithm did not return a solution within the prescribed time limit of two hours in any of the three runs. \label{tbl:aggregated_results_k5_ofv_cpu}}
\end{table}

For larger values of $K$, the \ABA~algorithm outperforms the \texttt{fast-anticlustering} algorithms not only in terms of speed, but also in terms of solution quality. The outperformance in terms of solution quality becomes more pronounced with larger values of $K$. Table~\ref{tbl:aggregated_results_k1000_ofv_cpu} presents the results for all datasets when $K=1{,}000$ and supports this conclusion.

\begin{table}
	\tiny
	\setlength{\tabcolsep}{4.7pt}
	\begin{tabular}{lrrrrrrrrrrrrr}
\toprule 
 \multicolumn{14}{l}{$K$= 1000} \\ \midrule
\multicolumn{4}{c}{} & \multicolumn{5}{c}{Deviation from ofv \textsf{ABA} [\%]} &  & \multicolumn{4}{c}{Deviation from cpu \textsf{ABA} [\%]} \\ \cmidrule(lr){5-9} \cmidrule(lr){11-14}
Dataset & $N$ & $D$ & ofv \textsf{ABA} & P-N5 & P-R5 & P-R50 & P-R500 & Rand & cpu \textsf{ABA} [s] & P-N5 & P-R5 & P-R50 & P-R500 \\
\cmidrule(lr){1-3} \cmidrule(lr){4-9} \cmidrule(lr){10-14}
Travel & 5,454 & 24 & \bfseries 126,560.01 & -11.0119 & -4.9807 & -2.4646 & -0.9294 & -15.3408 & \bfseries 0.201 & 345.7 & 244.1 & 353.3 & 1,302.7 \\
Npi & 10,440 & 40 & \bfseries 86,399.25 & -3.0555 & -2.9480 & -1.6556 & -0.9476 & -8.7383 & \bfseries 0.198 & 4,705.3 & 544.1 & 843.4 & 3,285.6 \\
Creditcard & 30,000 & 24 & \bfseries 718,236.40 & -1.0242 & -0.3735 & -0.1785 & -0.0928 & -3.0590 & \bfseries 1.432 & 441.1 & 163.8 & 254.7 & 862.0 \\
Adult & 32,561 & 110 & \bfseries 3,528,501.69 & -0.9517 & -0.2866 & -0.0912 & -0.0178 & -1.6242 & \bfseries 2.167 & 462.1 & 112.7 & 298.7 & 1,800.9 \\
Plants & 34,781 & 70 & \bfseries 257,738.24 & -1.5517 & -0.3786 & -0.1851 & -0.0914 & -2.6860 & \bfseries 0.965 & 1,735.2 & 413.1 & 694.9 & 2,761.4 \\
Bank & 45,211 & 53 & \bfseries 2,390,566.35 & -1.1955 & -0.2490 & -0.0754 & -0.0230 & -1.4508 & \bfseries 2.441 & 535.0 & 161.2 & 305.4 & 1,324.7 \\
Cifar10 & 50,000 & 3,072 & \bfseries 9,498,427.38 & -0.1982 & -0.2675 & -0.1422 & -0.0737 & -1.7665 & \bfseries 9.554 & 43,767.3 & 776.5 & 6,862.8 & 68,844.7 \\
Mnist & 60,000 & 784 & \bfseries 3,157,995.65 & -0.3111 & -0.2689 & -0.1400 & -0.0687 & -1.4934 & \bfseries 2.831 & 107,984.6 & 771.7 & 5,586.4 & 53,885.2 \\
Survival & 110,204 & 4 & 440,792.96 & -0.9112 & 0.0005 & \bfseries 0.0006 & \bfseries 0.0006 & -0.8635 & \bfseries 0.392 & 3,568.2 & 3,472.8 & 4,613.7 & 7,682.4 \\
Diabetes & 253,680 & 22 & 5,580,808.28 & -0.1361 & -0.0012 & -0.0001 & \bfseries 0.0002 & -0.3815 & \bfseries 1.864 & 6,697.1 & 1,846.1 & 2,530.4 & 6,665.0 \\
Music & 515,345 & 91 & \bfseries 46,896,196.66 & \NA & -0.0043 & -0.0015 & -0.0007 & -0.1929 & \bfseries 4.794 & \NA & 1,586.8 & 3,319.5 & 19,038.5 \\
Covtype & 581,012 & 55 & \bfseries 31,949,066.78 & -0.1218 & -0.0006 & \bfseries 0.0000 & \bfseries 0.0000 & -0.1296 & \bfseries 6.526 & 1,650.6 & 1,293.9 & 2,212.4 & 10,522.5 \\
Imagenet8 & 1,281,167 & 192 & \bfseries 13,327,165.86 & \NA & -0.0010 & -0.0005 & -0.0003 & -0.0794 & \bfseries 19.773 & \NA & 1,100.0 & 3,293.8 & 23,787.2 \\
Imagenet32 & 1,281,167 & 3,072 & \bfseries 265,670,329.59 & \NA & \NA & \NA & \NA & -0.0762 & \bfseries 481.641 & \NA & \NA & \NA & \NA \\
Census & 2,458,285 & 68 & \bfseries 167,163,196.65 & \NA & -0.0001 & \bfseries 0.0000 & \bfseries 0.0000 & -0.0400 & \bfseries 25.704 & \NA & 1,416.8 & 2,701.0 & 13,882.7 \\
Finance & 6,362,620 & 12 & \bfseries 76,350,416.57 & -0.0117 & \bfseries 0.0000 & \bfseries 0.0000 & \NA & -0.0141 & \bfseries 106.076 & 948.4 & 781.4 & 1,062.8 & \NA \\\cmidrule(lr){1-3} \cmidrule(lr){4-9} \cmidrule(lr){10-14} 
 Average & & &  & -1.7067 & -0.6506 & -0.3289 & -0.1603 & -2.3710 &  & 14,403.4 & 979.0 & 2,328.9 & 15,403.2 \\
\bottomrule \\ [-1.5ex] \multicolumn{7}{l}{\textsf{ABA} is the contribution in this paper}
\end{tabular}

	\caption{Comparing solution quality and running times for $K=1{,}000$ of \ABA, P-N5, P-R5, P-R50, P-R500, and random. Reported are the objective function values (ofv) for the \ABA~algorithm and the percentage deviations from these values for the other algorithms (positive deviation means improving on \ABA). Also reported are the running times of the \ABA~algorithm in seconds (cpu) and the percentage deviations from these values for the other algorithms (negative deviation means improving on \ABA). A dash indicates that the algorithm did not return a solution within the prescribed time limit of two hours in any of the three runs. \label{tbl:aggregated_results_k1000_ofv_cpu}}
\end{table}

A further analysis showed that the \ABA~algorithm produces solutions with high anticluster similarity, i.e., the anticlusters are very similar to each other in terms of diversity. We measured the diversity of an anticluster as the sum of squared distances between its assigned objects and its centroid. The benchmark approaches P-N5, P-R5, P-R50, P-R500, and Rand tend to produce anticlusters that are far less balanced in terms of diversity. To quantify the similarity of anticlusters within a solution, we compute, for each of the $K$ anticlusters, the diversity, i.e., the sum of squared distances between the assigned objects and the corresponding centroid. Based on these $K$ values, we then calculate two statistics, namely the standard deviation and the range (the difference between the maximum and minimum). Table~\ref{tbl:aggregated_results_k5_std_range} presents these statistics for all datasets and benchmark algorithms for $K=5$. Columns 4 and 10 report for each dataset the standard deviation (sd) and the range (range) obtained with \ABA~, respectively. The columns 5--9 provide for each benchmark algorithm the percentage deviations from the standard deviations of \ABA. Analogously, columns 11--15 provide for each benchmark algorithm the percentage deviations from the range values of \ABA. Positive deviations indicate that the corresponding benchmark algorithm produced a solution with a higher standard deviation or larger range. From Table~\ref{tbl:aggregated_results_k5_std_range}, we observe that the \ABA~algorithm consistently produces anticlusters with more balanced diversity. The same conclusion can be drawn for $K = \{2, 50, 500, 1{,}000, 2{,}000, 5{,}000\}$ as can be seen in the corresponding tables on \href{https://github.com/phil85/aba-results}{\color{blue}GitHub}. 

\begin{table}
	\tiny
	\setlength{\tabcolsep}{2.8pt}
	\begin{tabular}{lrrrrrrrrrrrrrr}
\toprule 
 \multicolumn{15}{l}{$K$= 5} \\ \midrule
\multicolumn{4}{c}{} & \multicolumn{5}{c}{Deviation from sd \textsf{ABA} [\%]} &  & \multicolumn{5}{c}{Deviation from range \textsf{ABA} [\%]} \\ \cmidrule(lr){5-9} \cmidrule(lr){11-15}
Dataset & $N$ & $D$ & sd \textsf{ABA} & P-N5 & P-R5 & P-R50 & P-R500 & Rand & range \textsf{ABA} [s] & P-N5 & P-R5 & P-R50 & P-R500 & Rand \\
\cmidrule(lr){1-3} \cmidrule(lr){4-9} \cmidrule(lr){10-15}
Travel & 5,454 & 24 & \bfseries 4.8 & 7,667.8 & 3,608.7 & 3,631.3 & 4,152.6 & 3,082.9 & \bfseries 13.3 & 7,821.7 & 3,518.5 & 3,689.9 & 4,207.4 & 3,332.3 \\
Npi & 10,440 & 40 & \bfseries 0.3 & 245.0 & 440.8 & 280.8 & 466.0 & 27,579.4 & \bfseries 0.9 & 226.3 & 406.2 & 282.4 & 354.1 & 26,261.8 \\
Creditcard & 30,000 & 24 & \bfseries 2,475.4 & 114.7 & 115.2 & 103.9 & 122.7 & 208.2 & \bfseries 6,985.7 & 102.2 & 114.5 & 88.0 & 125.3 & 202.9 \\
Adult & 32,561 & 110 & \bfseries 11,616.3 & 31.2 & 2.6 & 24.5 & 6.4 & 69.6 & \bfseries 29,830.0 & 35.7 & 14.3 & 33.0 & 13.6 & 84.5 \\
Plants & 34,781 & 70 & \bfseries 0.9 & 997.3 & 635.1 & 823.1 & 947.5 & 66,157.8 & \bfseries 2.9 & 932.1 & 597.4 & 727.5 & 919.6 & 62,448.5 \\
Bank & 45,211 & 53 & 5,303.9 & 11.4 & 7.9 & -2.4 & 1.7 & \bfseries -2.7 & \bfseries 13,524.9 & 17.8 & 14.8 & 1.4 & 5.6 & 4.7 \\
Cifar10 & 50,000 & 3,072 & \bfseries 38.4 & 19,311.3 & 21,843.0 & 27,201.3 & 20,001.0 & 23,785.2 & \bfseries 98.9 & 21,663.2 & 23,558.9 & 29,167.5 & 20,656.8 & 27,983.1 \\
Mnist & 60,000 & 784 & \bfseries 3.9 & 5,659.0 & 3,781.6 & 3,837.8 & 4,381.2 & 31,747.0 & \bfseries 10.6 & 5,838.0 & 3,671.2 & 3,822.8 & 4,250.1 & 31,659.9 \\
Survival & 110,204 & 4 & \bfseries 3.6 & 19,239.1 & 5,936.4 & 8,222.4 & 8,351.9 & 12,514.1 & \bfseries 9.5 & 18,843.9 & 6,215.4 & 8,744.6 & 9,263.3 & 13,148.7 \\
Diabetes & 253,680 & 22 & \bfseries 6.5 & 40,109.3 & 13,297.6 & 12,622.8 & 10,853.3 & 43,579.7 & \bfseries 15.6 & 45,253.3 & 15,983.7 & 15,023.4 & 12,706.3 & 44,789.9 \\
Music & 515,345 & 91 & \bfseries 5,872.7 & \NA & 322.2 & 408.4 & 452.0 & 794.0 & \bfseries 16,478.4 & \NA & 314.6 & 389.2 & 431.7 & 785.1 \\
Covtype & 581,012 & 55 & \bfseries 93,564.1 & 13.1 & 1.0 & 1.6 & 1.0 & 37.8 & \bfseries 192,502.9 & 40.4 & 3.0 & 2.8 & 2.0 & 68.1 \\
Imagenet8 & 1,281,167 & 192 & \bfseries 0.8 & \NA & 227,728.0 & 243,564.5 & 451,199.2 & 433,347.3 & \bfseries 2.3 & \NA & 247,409.1 & 242,311.0 & 482,542.0 & 454,750.8 \\
Imagenet32 & 1,281,167 & 3,072 & \bfseries 8.0 & \NA & \NA & \NA & \NA & 568,933.2 & \bfseries 21.9 & \NA & \NA & \NA & \NA & 608,410.1 \\
Census & 2,458,285 & 68 & \bfseries 68.2 & \NA & 4,940.4 & 5,782.9 & 6,532.9 & 50,867.9 & \bfseries 185.3 & \NA & 5,122.2 & 5,648.2 & 6,828.8 & 56,229.6 \\
Finance & 6,362,620 & 12 & \bfseries 151,605.7 & 289.7 & 34.0 & 18.8 & \NA & 341.9 & \bfseries 382,057.0 & 338.4 & 47.4 & 28.8 & \NA & 394.8 \\\cmidrule(lr){1-3} \cmidrule(lr){4-9} \cmidrule(lr){10-15} 
 Average & & &  & 7,807.4 & 18,846.3 & 20,434.8 & 36,247.8 & 78,940.2 &  & 8,426.1 & 20,466.1 & 20,664.0 & 38,736.2 & 83,159.7 \\
\bottomrule \\ [-1.5ex] \multicolumn{7}{l}{\textsf{ABA} is the contribution in this paper}
\end{tabular}

	\caption{Comparing the statistics of the solutions produced by \ABA, P-N5, P-R5, P-R50, P-R500 and random. Reported are the standard deviations (sd) of the diversity values of the $K=5$ anticlusters produced by the \ABA~algorithm and the percentage deviations from these values for the other algorithms (negative deviation means improving on \ABA). Also reported are the ranges (range) of the diversity values of the $K=5$ anticlusters (max diversity - min diversity) produced by the \ABA~algorithm and the percentage deviations from these values for the other algorithms (negative deviation means improving on \ABA). A dash indicates that the algorithm did not return a solution within the prescribed time limit of two hours. \label{tbl:aggregated_results_k5_std_range}}
\end{table}

\begin{figure}
	\centering
	\includegraphics[width=0.49\textwidth]{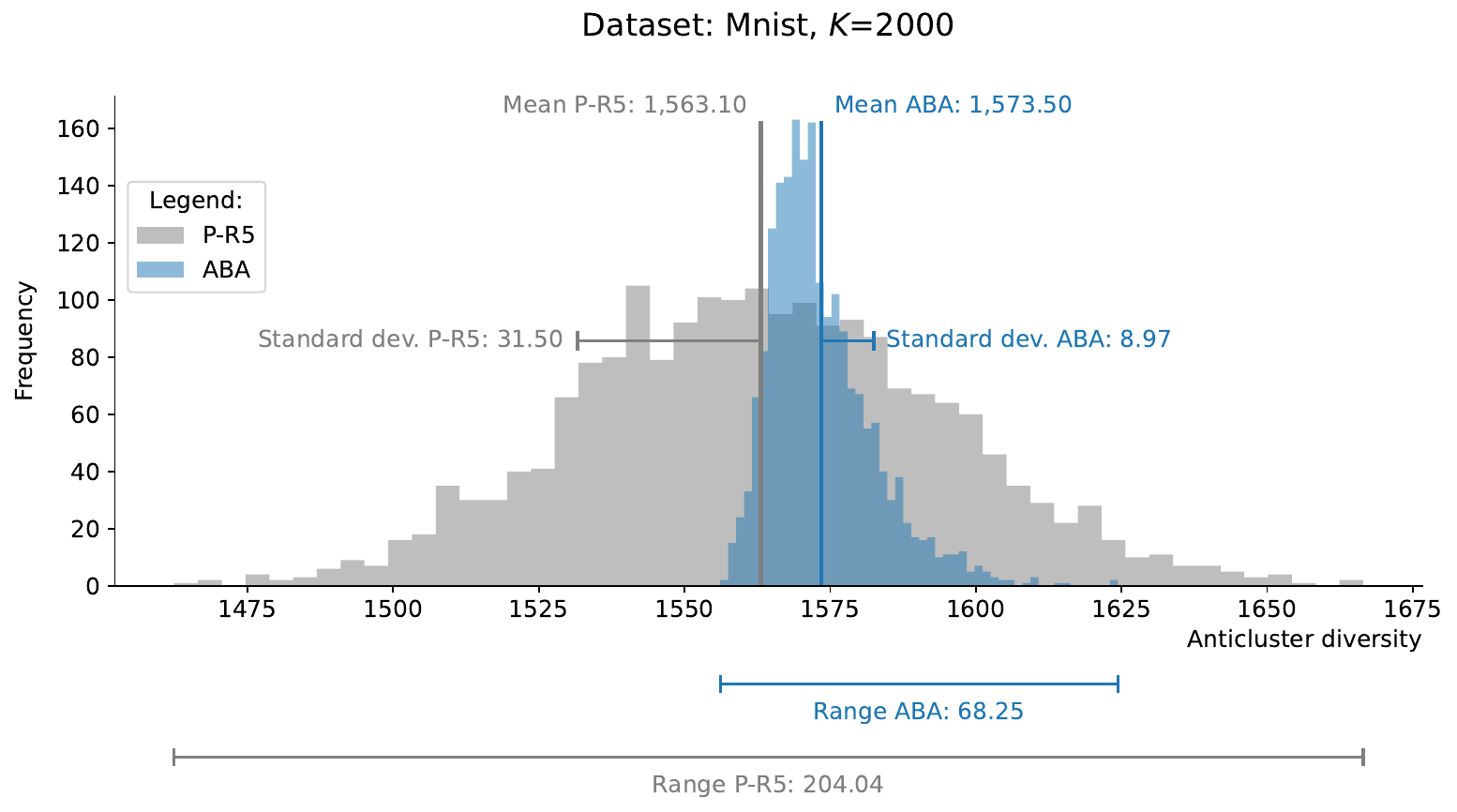}
	\includegraphics[width=0.49\textwidth]{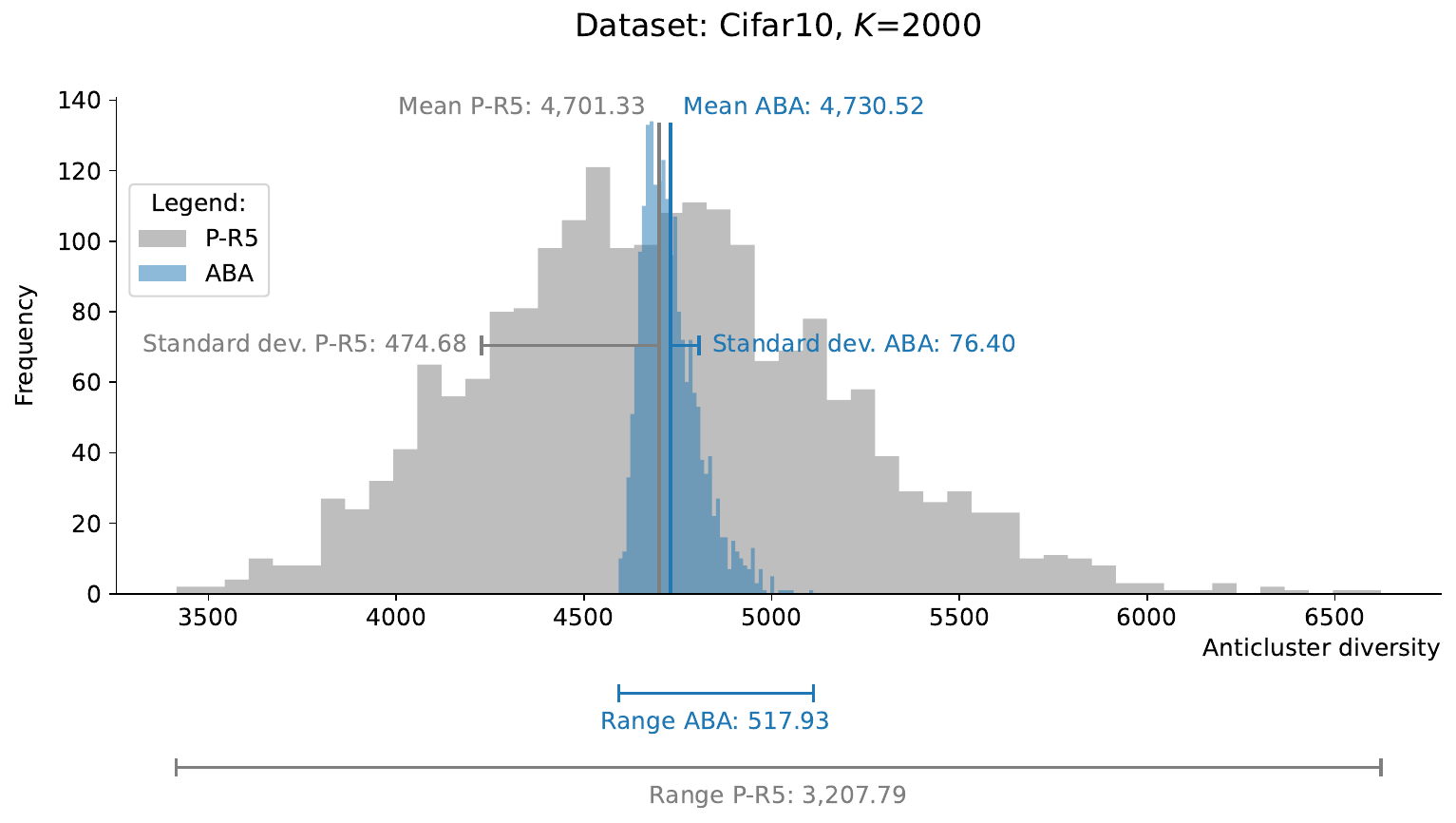}
	\includegraphics[width=0.49\textwidth]{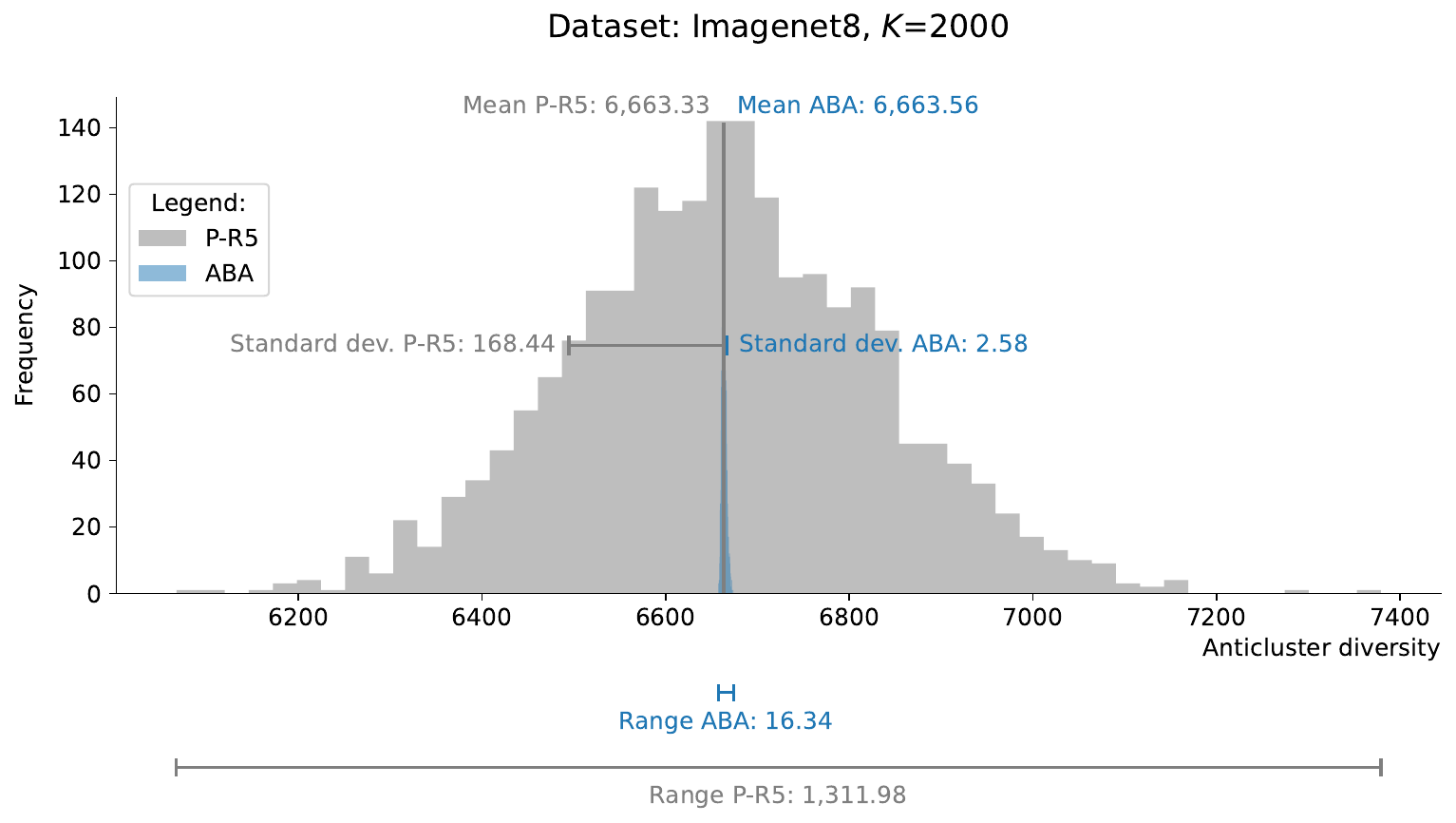}
	\includegraphics[width=0.49\textwidth]{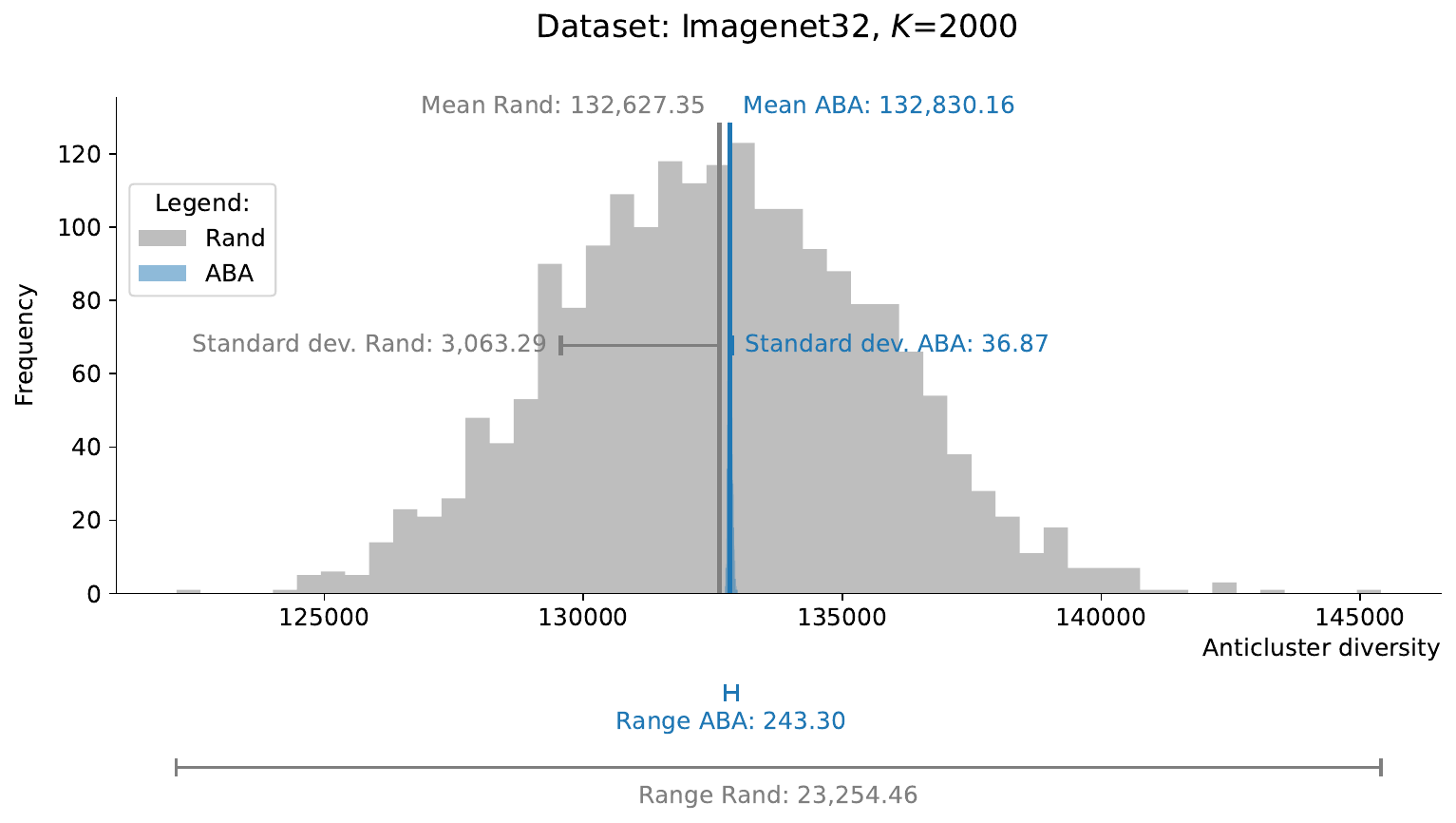}
	\caption{Comparison of distribution of anticluster diversity values between \ABA~and P-R5 for $K=2{,}000$. For the dataset \texttt{Imagenet32}, the P-R5 algorithm found no solution within the time limit of two hours. We therefore compare \ABA~and Rand for this dataset. \label{fig:distribution1}}
\end{figure}

We observed the largest differences in terms of balanced diversity between the \ABA~algorithm and the benchmark algorithms for the image datasets \texttt{Mnist}, \texttt{Cifar10}, \texttt{Imagenet8}, and \texttt{Imagenet32} when $K$ is large. Figure~\ref{fig:distribution1} shows the distributions of the anticluster diversity values for the solutions obtained with \ABA~and P-R5 with $K=2{,}000$. For the dataset \texttt{Imagenet32}, the P-R5 algorithm found no solution within the time limit of two hours. We therefore compare \ABA~and Rand for this dataset. The diversity distributions of \ABA~exhibit not only higher means, but also smaller spreads than the diversity distributions of P-R5 and Rand. 

\begin{figure}[ht]
	\includegraphics[width=\textwidth]{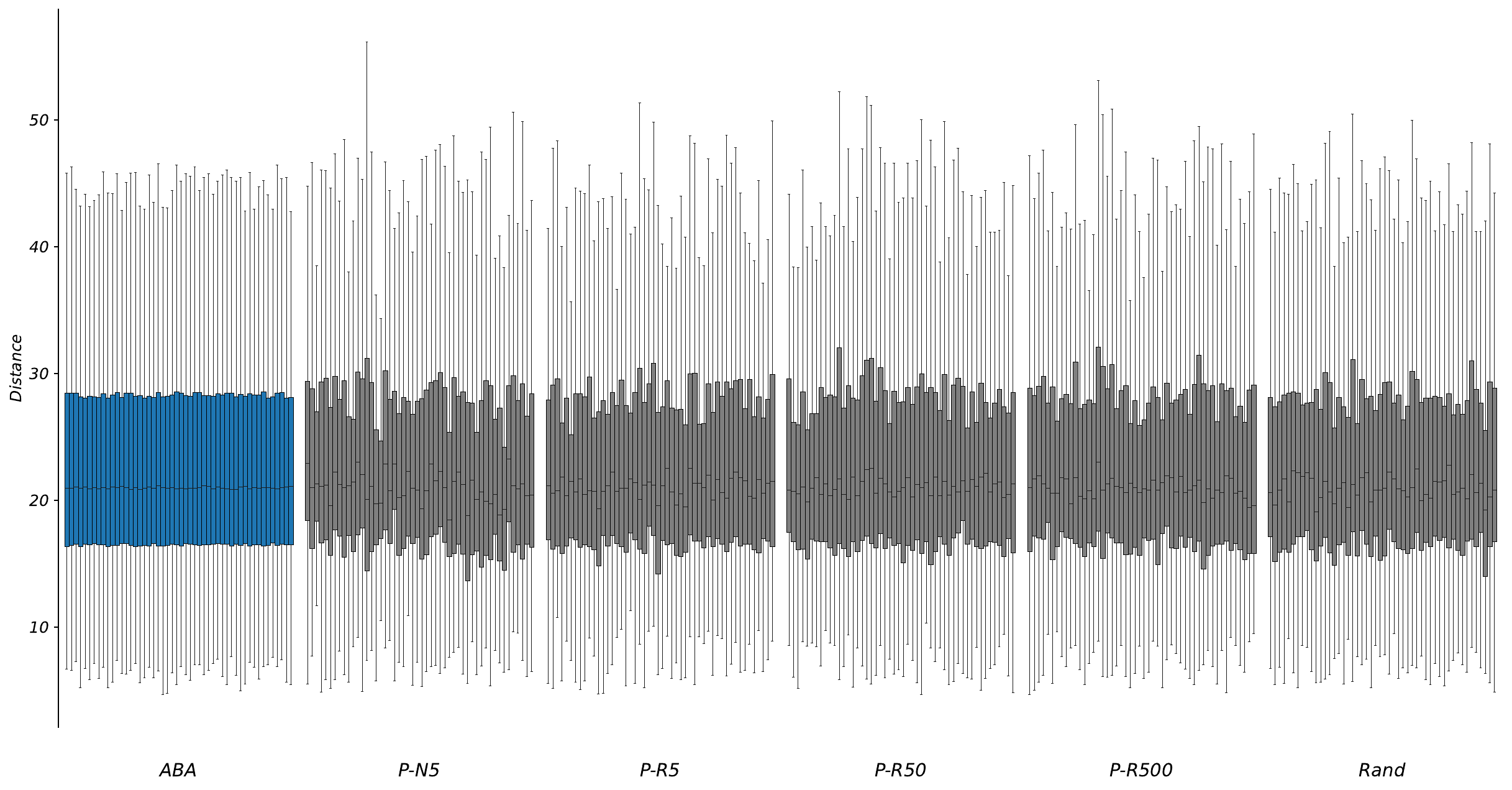}
	\caption{Comparison of distance distributions within anticlusters for the \texttt{Travel} dataset with $K=50$. For each algorithm, 50 boxplots are shown. Each boxplot shows the distribution of distances between the objects and the centroid of an anticluster. The outliers have been removed for clarity.\label{fig:boxplots}}
\end{figure}

To further illustrate the balanced diversity of anticlusters produced by \ABA, we looked at the distribution of distances within anticlusters. Figure~\ref{fig:boxplots} visualizes for the \texttt{Travel} dataset with $K=50$ the distribution of distances within anticlusters for the solutions obtained by \ABA~and the benchmark algorithms. For each algorithm, 50 boxplots are shown, one for each anticluster. Each boxplot shows the distribution of distances between the objects and the centroid of the anticluster. The outliers have been removed for clarity. We can see that the \ABA~algorithm produces anticlusters with very similar distance distributions, whereas the benchmark algorithms yield anticlusters with highly varying distance distributions. The \ABA~algorithm produces anticlusters with similar distance distribution because it first sorts all objects by their distance to the global centroid and then splits them into batches of size $K$. This ensures that the objects in each batch have similar distances to the global centroid and since anticluster centroids tend to be close to the global centroid in final solutions, each anticluster gets one object from each distance range. 

\subsection{Effectiveness of the Hierarchical Decomposition Strategy}\label{experiment:hierarchical_decomposition}

\begin{figure}
	\centering
	\includegraphics[width=0.8\textwidth]{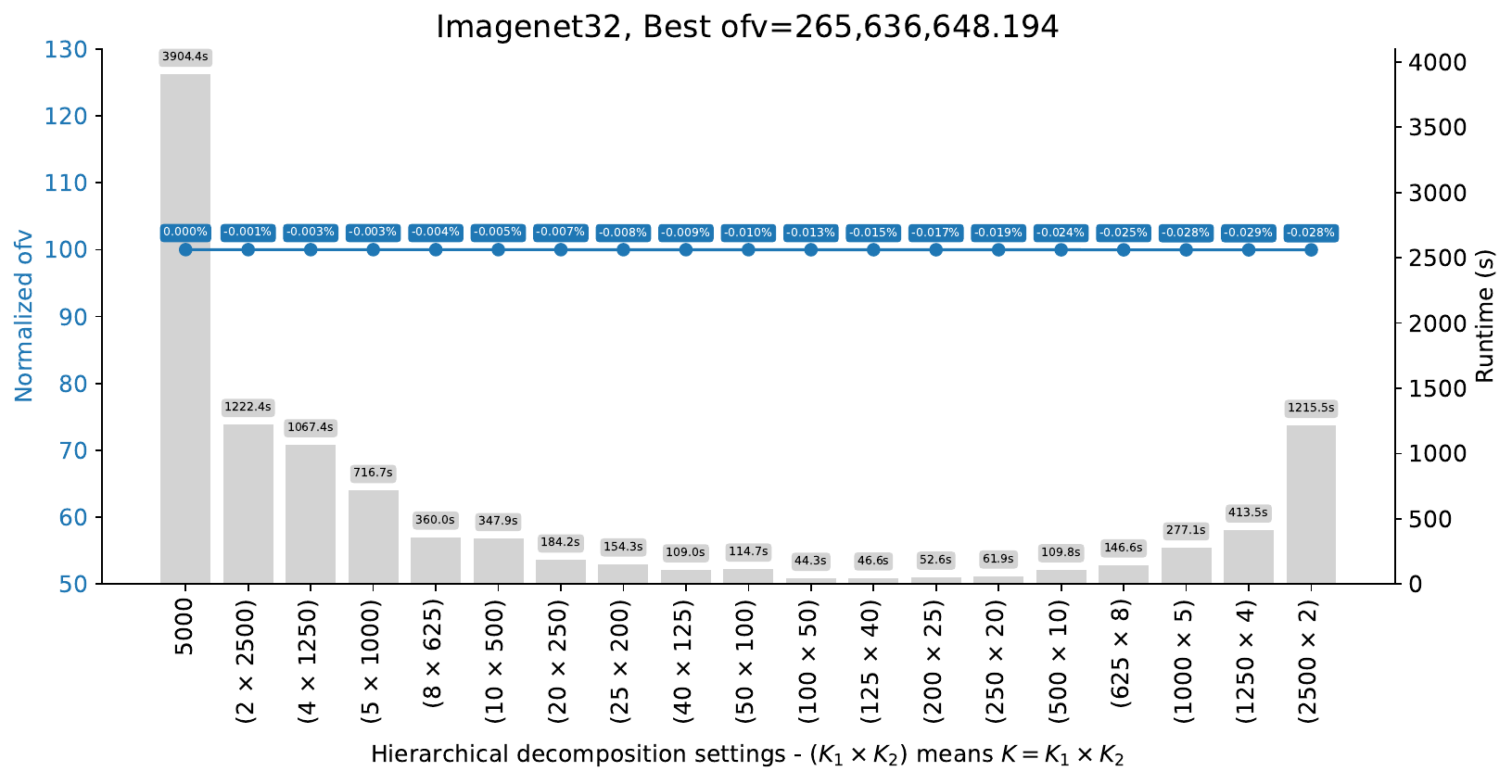}
	\caption{Comparison of objective function values (ofv) and running times (cpu) for different hierarchical decomposition strategies. Each setting produces $K=5,000$ anticlusters for the Imagenet32 dataset. The blue line shows the normalized objective function values for the different settings with labels indicating the percentage deviation from the best setting (see left axis). The gray bars represent the running time in seconds for the different settings (see right axis). \label{fig:barplot}}
\end{figure}

The hierarchical decomposition strategy presented in Section~\ref{sec:algorithm:hierarchical} allows the \ABA~algorithm to compute high-quality solutions for large values of $K$ in short running time. To illustrate the effectiveness of this strategy, we applied different hierarchical settings on the \texttt{Imagenet32} dataset which has $N=1{,}281{,}167$ objects and $D=3{,}072$ features for $K=5{,}000$. Figure~\ref{fig:barplot} compares the running times and normalized objective function values for different two-level hierarchical strategies to the base variant of \ABA~that uses no hierarchical decomposition. With no hierarchical decomposition, the \ABA~algorithm requires more than 3,900 seconds to compute a solution. By applying hierarchical decomposition, the running time can be drastically reduced to 44.3 seconds with only a marginal deterioration in the objective function value (-0.013\%). As observed experimentally in Figure~\ref{fig:barplot}, the quality of the solution is not noticeably affected by the selected decomposition of $K$. However, the running times are drastically improved by choosing a decomposition where the values of $K_\ell$ are as balanced as possible.  

With hierarchical decomposition, the \ABA~algorithm can handle even much larger values of $K$. To demonstrate this capability, we applied \ABA~with $K = \{10{,}000,\allowbreak 20{,}000,\allowbreak 40{,}000,\allowbreak 
80{,}000,\allowbreak 160{,}000,\allowbreak 320{,}000,\allowbreak 640{,}000\}$ to the \texttt{Imagenet32} dataset which has $N=1{,}281{,}167$ objects. Table~\ref{tbl:table_large_k_imagenet32} lists the running times and the objective function values for \ABA. For these instances, we applied the \ABA~algorithm with the hierarchical decomposition settings provided in Table~\ref{tbl:hierarchical_large}. The only benchmark approach that was able to produce solutions within a time limit of two hours was random partitioning (Rand). We can see from Table~\ref{tbl:table_large_k_imagenet32} that \ABA~clearly outperforms Rand in terms of solution quality for all considered values of $K$. The absolute percentage deviation of the objective function values of Rand from those of \ABA~increases with larger values of $K$. The \ABA~algorithm obtains solutions that are over 30\% better than those of Rand for $K=640{,}000$. 

\begin{table}
	\centering
	\small
	\begin{tabular}{lrrrrrrr}
		& \multicolumn{7}{c}{$K$} \\ \cmidrule(lr){2-8}
		& 10{,}000 & 20{,}000 & 40{,}000 & 80{,}000 & $160{,}000$ & $320{,}000$ & $640{,}000$ \\ \midrule
		\texttt{Imagenet32} & (50$\times$200) & (100$\times$200) & (200$\times$200) & (2$\times$200$\times$200) & (4$\times$200$\times$200) & (8$\times$200$\times$200) & (16$\times$200$\times$200) \\
	\end{tabular}
	\caption{The hierarchical decomposition settings used for the \texttt{Imagenet32} dataset as a function of the number of anticlusters ($K$). The expression $(K_1 \times K_2 \times K_3)$ denotes a decomposition with three levels and $K=K_1 \times K_2 \times K_3$.\label{tbl:hierarchical_large}}
\end{table}

\begin{table}
	\scriptsize
	\setlength{\tabcolsep}{16pt}
	\begin{tabular}{rrrrrrr}
\toprule 
 \multicolumn{7}{l}{Dataset: Imagenet32 ($N$=1,281,167, $D$=3,072)} \\ \midrule
$K$ & Min size & Max size & cpu \textsf{ABA} [s] & ofv \textsf{ABA} & ofv Rand & Deviation [\%] \\
\midrule
10,000 & 128 & 129 & 87.9 & \bfseries 265,506,874.2 & 263,596,017.4 & -0.7197 \\
20,000 & 64 & 65 & 88.5 & \bfseries 265,211,753.5 & 261,523,122.1 & -1.3908 \\
40,000 & 32 & 33 & 106.4 & \bfseries 264,440,895.2 & 257,383,295.3 & -2.6689 \\
80,000 & 16 & 17 & 138.7 & \bfseries 262,466,218.2 & 249,093,874.6 & -5.0949 \\
160,000 & 8 & 9 & 175.0 & \bfseries 257,433,000.0 & 232,437,686.9 & -9.7094 \\
320,000 & 4 & 5 & 254.9 & \bfseries 244,033,085.9 & 199,315,726.3 & -18.3243 \\
640,000 & 2 & 3 & 456.5 & \bfseries 191,492,745.8 & 132,899,446.9 & -30.5982 \\
\bottomrule \\ [-1.5ex] \multicolumn{7}{l}{\textsf{ABA} is the contribution in this paper}
\end{tabular}

	\caption{Comparing solution quality of \ABA~and Rand on the Imagenet32 dataset for large values of $K$. Reported are the objective function values (OFV) of both algorithms as well as the percentage deviation of the Rand values from the \ABA~values. Columns 2 and 3 state the minimum and maximum number of objects in an anticluster and column 5 reports the running time of \ABA. The algorithms P-N5, P-R5, P-R50, and P-R500 were not able to find solutions for such large values of $k$.  \label{tbl:table_large_k_imagenet32}}
\end{table}

\subsection{Application of \ABA~to Balanced \texorpdfstring{$K$-cut}{K-Cut}}\label{experiment:balanced_cut}

Recall that the balanced $K$-cut problem consists of partitioning the nodes of a graph into $K$ equal-sized anticlusters such that the sum of edge weights between the anticlusters (the cut cost) is minimized. 
In the case of tabular data, where the underlying graph is complete and the edge weights represent squared Euclidean distances between the corresponding objects (or data points), minimizing the cut cost corresponds to maximizing the within-anticluster sum of squared distances. Because of that equivalence, we can use the \ABA~algorithm for the balanced $K$-cut problem of tabular data. In this section, we compare the \ABA~algorithm to \METIS, the state-of-the-art algorithm for solving balanced $K$-cut problems. The \METIS\ algorithm is applicable to a more general setup than Euclidean anticlustering -- to any general graph. It takes as input the graph representation in the form of an adjacency list with edge weights. 
As a result applying \METIS\ to Euclidean anticlustering requires to create a representation of a complete graph. Another requirement of \METIS\ is that edge weights are integer. In order to address that, we multiply the squared Euclidean distances by a factor of $1{,}000$ and rounded the resulting values to the nearest integer. 


We apply \METIS\ here with its default parameter settings, except that we set the parameter \texttt{ufactor} to its minimum value of one. The parameter \texttt{ufactor} controls the maximum allowed deviation of anticluster sizes from the target size ($N/K$). Specifically, a value of $x$ for parameter \texttt{ufactor} allows deviations of up to $x/1000$, where $x$ must be larger or equal to one and integer. For small values of \texttt{ufactor}, we found that \METIS\ occasionally fails to return a solution. In such cases, we increased \texttt{ufactor} by one and reran \METIS. This process was repeated until a solution was found or a time limit of $7{,}200$ seconds was reached. Because constructing the graph representation for \METIS\ and running \METIS\ on it is computationally expensive and memory-intensive, we limited this experiment to datasets with at most $30{,}000$ objects from Table~\ref{tbl:datasets} and Table~\ref{tbl:datasets_categories}. For datasets with fewer than $10{,}000$ objects, we tested $K=\{2, 5, 50, 500, 1{,}000, 2{,}000\}$. For datasets with at least $10{,}000$ objects, we additionally tested $K=5{,}000$. Random partitioning (Rand) again served as a baseline. \ABA\ is applied in this experiment without the hierarchical decomposition strategy. 

Table~\ref{tbl:ofv_cpu_table_with_metis} reports the results. Note that for several combinations of datasets and values of $K$, \METIS\ returned anticlusters whose sizes differed by more than one object, even when \texttt{ufactor} was set to its minimum value. Because comparing objective values is only meaningful when the anticluster size constraints are satisfied, these instances are excluded from Table~\ref{tbl:ofv_cpu_table_with_metis} (the complete results are provided in the Electronic Companion to this paper on \href{https://github.com/phil85/aba-results}{\color{blue}GitHub}). The first four columns of Table~\ref{tbl:ofv_cpu_table_with_metis} list the dataset name, the number of objects ($N$), the number of features ($D$), and the number of anticlusters ($K$). Column~5 reports the within-anticluster sum of squared distances ($W(\mathcal{C})$) obtained by \ABA. Columns~6 and 7 report the percentage deviation of the $W(\mathcal{C})$ values obtained by the benchmark algorithms from that of \ABA. Negative values indicate that \ABA\ performs better. Columns~8 and 9 give the total running time of \ABA\ and \METIS\ in seconds, respectively. Column~9 lists separately the running time required to construct the \METIS\ input graph, i.e., the adjacency list. This running time is included in the total running time reported in column 9. Columns~10 and 11 report the ratio between the sizes of the smallest and largest anticlusters for \ABA\ and \METIS, respectively. This ratio equals 100\% when all anticlusters have the same size. If $N$ is not divisible by $K$, we still report 100\% whenever the largest and smallest anticlusters differ in size by at most one object. The best values for each instance, are highlighted in bold and the last row reports average values. The results in Table~\ref{tbl:ofv_cpu_table_with_metis} demonstrate that \ABA~produces better solutions than \METIS\ for all but one instance (Facebook with $K=2{,}000$). 
The outperformance of \ABA\ generally increases with increasing values of $K$, consistent with our observations on the poor quality of other algorithms for small anticlusters. \ABA\ also consistently outperforms random partitioning (Rand), and often by a substantial margin. In addition, \ABA\ is considerably faster than \METIS, for some instances by several orders of magnitude. 

\begin{table}
	\scriptsize
	\setlength{\tabcolsep}{6.5pt}
	\begin{tabular}{lrrrrrrrrrrr}
\toprule
\multicolumn{5}{c}{} & \multicolumn{2}{c}{Deviation [\%]} & \multicolumn{3}{c}{cpu [s]} & \multicolumn{2}{c}{min/max ratio [\%]} \\ \cmidrule(lr){6-7} \cmidrule(lr){8-10} \cmidrule(lr){11-12}
Dataset & $N$ & $D$ & $K$ & $W(\mathcal{C})$ \textsf{ABA} & METIS & Rand & \textsf{ABA} & METIS & METIS input & \textsf{ABA} & METIS \\
\midrule
Abalone & 4,177 & 10 & 50 & \bfseries 3,488,723.9 & -0.035 & -0.887 & \bfseries 0.047 & 5.27 & 4.81 & 100 & 100 \\
Abalone & 4,177 & 10 & 1000 & \bfseries 174,637.7 & -14.806 & -23.031 & \bfseries 1.234 & 10.49 & 4.81 & 100 & 100 \\
Travel & 5,454 & 24 & 5 & \bfseries 142,755,047.4 & -0.009 & -0.019 & \bfseries 0.010 & 8.58 & 8.02 & 100 & 100 \\
Travel & 5,454 & 24 & 50 & \bfseries 14,274,801.6 & -0.409 & -0.457 & \bfseries 0.038 & 8.71 & 8.02 & 100 & 100 \\
Travel & 5,454 & 24 & 500 & \bfseries 1,420,006.5 & -1.689 & -8.061 & \bfseries 0.444 & 13.69 & 8.02 & 100 & 100 \\
Travel & 5,454 & 24 & 2000 & \bfseries 320,934.3 & -2.572 & -26.403 & \bfseries 3.427 & 32.56 & 8.02 & 100 & 100 \\
Facebook & 7,050 & 13 & 50 & \bfseries 12,918,006.7 & -0.972 & -0.484 & \bfseries 0.047 & 14.34 & 13.17 & 100 & 100 \\
Facebook & 7,050 & 13 & 2000 & 288,196.0 & \bfseries 1.657 & -15.150 & \bfseries 10.922 & 46.82 & 13.17 & 100 & 100 \\
Frogs & 7,195 & 22 & 50 & \bfseries 22,774,595.0 & -0.606 & -0.189 & \bfseries 0.041 & 14.88 & 13.69 & 100 & 100 \\
Frogs & 7,195 & 22 & 500 & \bfseries 2,275,447.5 & -3.800 & -3.705 & \bfseries 0.591 & 21.27 & 13.69 & 100 & 100 \\
Frogs & 7,195 & 22 & 1000 & \bfseries 1,132,930.0 & -8.065 & -8.410 & \bfseries 1.354 & 27.45 & 13.69 & 100 & 100 \\
Frogs & 7,195 & 22 & 2000 & \bfseries 549,425.0 & -13.112 & -18.914 & \bfseries 4.477 & 39.06 & 13.69 & 100 & 100 \\
Electric & 10,000 & 12 & 50 & \bfseries 23,997,442.2 & -0.038 & -0.497 & \bfseries 0.022 & 27.78 & 25.89 & 100 & 100 \\
Electric & 10,000 & 12 & 500 & \bfseries 2,398,509.8 & -1.245 & -4.960 & \bfseries 0.314 & 43.47 & 25.89 & 100 & 100 \\
Electric & 10,000 & 12 & 1000 & \bfseries 1,197,376.1 & -4.438 & -9.795 & \bfseries 0.731 & 53.17 & 25.89 & 100 & 100 \\
Electric & 10,000 & 12 & 2000 & \bfseries 594,629.9 & -9.326 & -19.524 & \bfseries 2.037 & 70.76 & 25.89 & 100 & 100 \\
Electric & 10,000 & 12 & 5000 & \bfseries 222,968.1 & -40.960 & -46.422 & \bfseries 8.207 & 183.24 & 25.89 & 100 & 100 \\
Npi & 10,440 & 40 & 50 & \bfseries 18,197,315.0 & -0.021 & -0.470 & \bfseries 0.036 & 31.69 & 28.79 & 100 & 100 \\
Npi & 10,440 & 40 & 500 & \bfseries 1,816,448.0 & -0.272 & -4.561 & \bfseries 0.192 & 57.39 & 28.79 & 100 & 100 \\
Pulsar & 17,898 & 8 & 50 & \bfseries 51,251,273.6 & -0.075 & -0.266 & \bfseries 0.120 & 91.23 & 84.19 & 100 & 100 \\
Pulsar & 17,898 & 8 & 500 & \bfseries 5,127,391.9 & -0.134 & -2.800 & \bfseries 2.103 & 143.11 & 84.19 & 100 & 100 \\
Pulsar & 17,898 & 8 & 1000 & \bfseries 2,564,076.7 & -0.700 & -5.506 & \bfseries 4.578 & 191.41 & 84.19 & 100 & 100 \\
Pulsar & 17,898 & 8 & 2000 & \bfseries 1,271,605.7 & -2.339 & -10.410 & \bfseries 14.390 & 298.74 & 84.19 & 100 & 100 \\
Pulsar & 17,898 & 8 & 5000 & \bfseries 512,941.9 & -2.107 & -26.454 & \bfseries 151.112 & 638.28 & 84.19 & 100 & 100 \\
Creditcard & 30,000 & 24 & 500 & \bfseries 43,179,045.4 & -1.830 & -1.608 & \bfseries 2.644 & 269.13 & 241.69 & 100 & 100 \\
Creditcard & 30,000 & 24 & 1000 & \bfseries 21,549,345.5 & -1.965 & -3.069 & \bfseries 5.949 & 342.82 & 241.69 & 100 & 100 \\
Creditcard & 30,000 & 24 & 2000 & \bfseries 10,703,564.1 & -3.938 & -5.837 & \bfseries 18.232 & 451.17 & 241.69 & 100 & 100 \\
Creditcard & 30,000 & 24 & 5000 & \bfseries 4,141,666.7 & -8.269 & -13.018 & \bfseries 135.097 & 880.22 & 241.69 & 100 & 100 \\\cmidrule(lr){1-4} \cmidrule(lr){5-7} \cmidrule(lr){8-10} \cmidrule(lr){11-12} 
 Average & & & &  & -4.360 & -9.318 & 13.157 & 143.45 & 60.63 & 100 & 100 \\
\bottomrule \\ [-1.5ex] \multicolumn{7}{l}{\textsf{ABA} is the contribution in this paper}
\end{tabular}

	\caption{Comparing solution quality and running times of \ABA, \METIS, and random (Rand). \label{tbl:ofv_cpu_table_with_metis}}
\end{table}

\section{Conclusions}\label{sec:conclusions}
In this paper, we present a new heuristic algorithm called Assignment-Based Anticlustering (\ABA), for solving large-scale Euclidean anticlustering problems. \ABA\ is scalable to an unprecedented degree due to its efficient exploitation of the properties of the Euclidean sum of square metric. 
On large-scale benchmark instances, \ABA\ consistently outperforms existing methods in both solution quality and running time, often by large margins, especially when the number of anticlusters is large. \ABA\ not only achieves the best objective function values and the fastest running times known for these datasets, but it is also superior to existing methods in terms of anticluster similarity, producing anticlusters with highly similar statistical properties. High anticluster similarity is recognized in the literature as a desirable property in practical anticlustering applications.

This work suggests several promising directions for future research. The basic concept of \ABA\ of pre-ranking objects and then assigning them in ranked order could be promising in other scenarios. For example, for single cluster diversity problems such as the maximum diversity problem (MDP, see \citealt{hochbaum2023breakpoints}) or for anticlustering problems with predefined, but not equal, anticluster sizes, and anticlustering that respect must-link and cannot-link constraints on the objects. 

\bibliographystyle{plainnat}
\bibliography{literature}

\pagebreak

\begin{section}*{Funding}
    The research of Dorit S.\ Hochbaum is supported in part by AI institute NSF award 2112533.    
\end{section}

\begin{section}*{Appendix}

\noindent We prove Fact~\ref{fact1}. Let $\mathcal{C}_k$ be an anticluster of size
$N_k = |\mathcal{C}_k|$ with centroid
\[
\bm{\mu}_k = \frac{1}{N_k}\sum_{i \in \mathcal{C}_k} \bm{x}_i .
\]
Using
\[
\|\bm{x}_i-\bm{x}_{i'}\|_2^2
= \|\bm{x}_i\|_2^2+\|\bm{x}_{i'}\|_2^2
  -2\bm{x}_i^\top\bm{x}_{i'},
\]
we obtain
\begin{align*}
\sum_{\substack{i,i' \in \mathcal{C}_k\\ i<i'}}
\|\bm{x}_i-\bm{x}_{i'}\|_2^2
&= (N_k-1)\sum_{i \in \mathcal{C}_k}\|\bm{x}_i\|_2^2
   -2\sum_{\substack{i,i' \in \mathcal{C}_k\\ i<i'}}
   \bm{x}_i^\top\bm{x}_{i'}  \\
&= N_k\sum_{i \in \mathcal{C}_k}\|\bm{x}_i\|_2^2
   - \Big\|\sum_{i \in \mathcal{C}_k}\bm{x}_i\Big\|_2^2 \\
&= N_k\sum_{i \in \mathcal{C}_k}\|\bm{x}_i\|_2^2
   - N_k^2\|\bm{\mu}_k\|_2^2 .
\end{align*}
On the other hand,
\begin{align*}
N_k\sum_{i \in \mathcal{C}_k}\|\bm{x}_i-\bm{\mu}_k\|_2^2
&= N_k\sum_{i \in \mathcal{C}_k}
   \left(\|\bm{x}_i\|_2^2
   -2\bm{x}_i^\top\bm{\mu}_k
   +\|\bm{\mu}_k\|_2^2\right) \\
&= N_k\sum_{i \in \mathcal{C}_k}\|\bm{x}_i\|_2^2
   - N_k^2\|\bm{\mu}_k\|_2^2 ,
\end{align*}
where we used $\sum_{i \in \mathcal{C}_k}\bm{x}_i=N_k\bm{\mu}_k$.
Thus,
\[
\sum_{\substack{i,i' \in \mathcal{C}_k\\ i<i'}}
\|\bm{x}_i-\bm{x}_{i'}\|_2^2
=
N_k\sum_{i \in \mathcal{C}_k}
\|\bm{x}_i-\bm{\mu}_k\|_2^2 .
\qquad \square
\]

\end{section}
\end{document}